\documentclass{article} 
\usepackage{iclr2027_conference,times}

\usepackage{amsmath,amsfonts,bm}

\def\eqref#1{equation~\ref{#1}}

\def\1{\bm{1}}

\DeclareMathAlphabet{\mathsfit}{\encodingdefault}{\sfdefault}{m}{sl}
\SetMathAlphabet{\mathsfit}{bold}{\encodingdefault}{\sfdefault}{bx}{n}

\usepackage{hyperref}
\usepackage{url}

\usepackage{graphicx}
\usepackage{booktabs}
\usepackage{multirow}
\usepackage[table]{xcolor}
\usepackage{wrapfig}
\usepackage{caption}
\usepackage{graphicx}
\usepackage{subcaption} 
\usepackage{amsmath,amssymb}
\usepackage{algorithm}
\usepackage{algorithmic}
\usepackage[T1]{fontenc}

\title{HyperErase: Scale-Calibrated Hypernetwork for Multi-Concept Erasure in Text-to-Image Models}

\author{%
Yi Sun$^{1}$ \quad Xinhao Zhong$^1$ \quad Zhiqi Zhang$^{3}$ \quad Yimin Zhou$^{2}$ \quad Junhao Li$^1$ \quad Yuxia Qiao$^5$\\
\bfseries  Bin Chen$^{1,4}$\thanks{Corresponding Author.}\\
$^1$Harbin Institute of Technology, Shenzhen \\ 
$^2$Tsinghua Shenzhen International Graduate School, Tsinghua University \\
$^3$Jilin University \\
$^4$Peng Cheng Laboratory \\
$^5$South China University of Technology\\
}

\iclrfinalcopy 
\begin{document}

\maketitle

\begin{abstract}

Recent advances in text-to-image (T2I) generation have substantially improved visual synthesis, but have also raised increasing safety concerns due to their potential to generate harmful or undesirable content. Existing concept erasure methods predominantly follow a static weight paradigm, producing a single frozen adapter that struggles to adapt to diverse prompt variations and suffers from parameter interference when scaling to multiple concepts. We propose \textbf{HyperErase}, a framework for concept erasure based on hypernetwork-driven prompt-conditioned parameter synthesis. Our approach first reframes concept erasure as prompt-conditioned parameter amortization and trains a hypernetwork to map textual descriptions to prompt-specific LoRA updates, eliminating the need for per-prompt gradient optimization or manual LoRA merging. To further improve the stability and precision of synthesized adapters, we develop a decoupled rectification strategy, which disentangles LoRA tokens into pattern and scale subspaces, applies a square-root transform to curb multiplicative over-scaling, and leverages teacher-derived canonical priors for inference-time correction. Extensive experiments across major concept categories demonstrate that HyperErase consistently improves the trade-off between erasure effectiveness, image quality, and semantic alignment, achieving performance comparable to gold-standard single-concept baselines. Furthermore, the resulting models can provide specialized LoRAs for each input prompt variation in a single forward pass without requiring gradient updates during inference. These principled and flexible framework offers a new paradigm for concept erasure in T2I models.

\end{abstract}

\section{Introduction}
\label{sec:introduction}

\begin{wrapfigure}{r}{0.5\textwidth}
  \vspace{-1em}
  \centering
  \resizebox{\linewidth}{!}{%
  \includegraphics{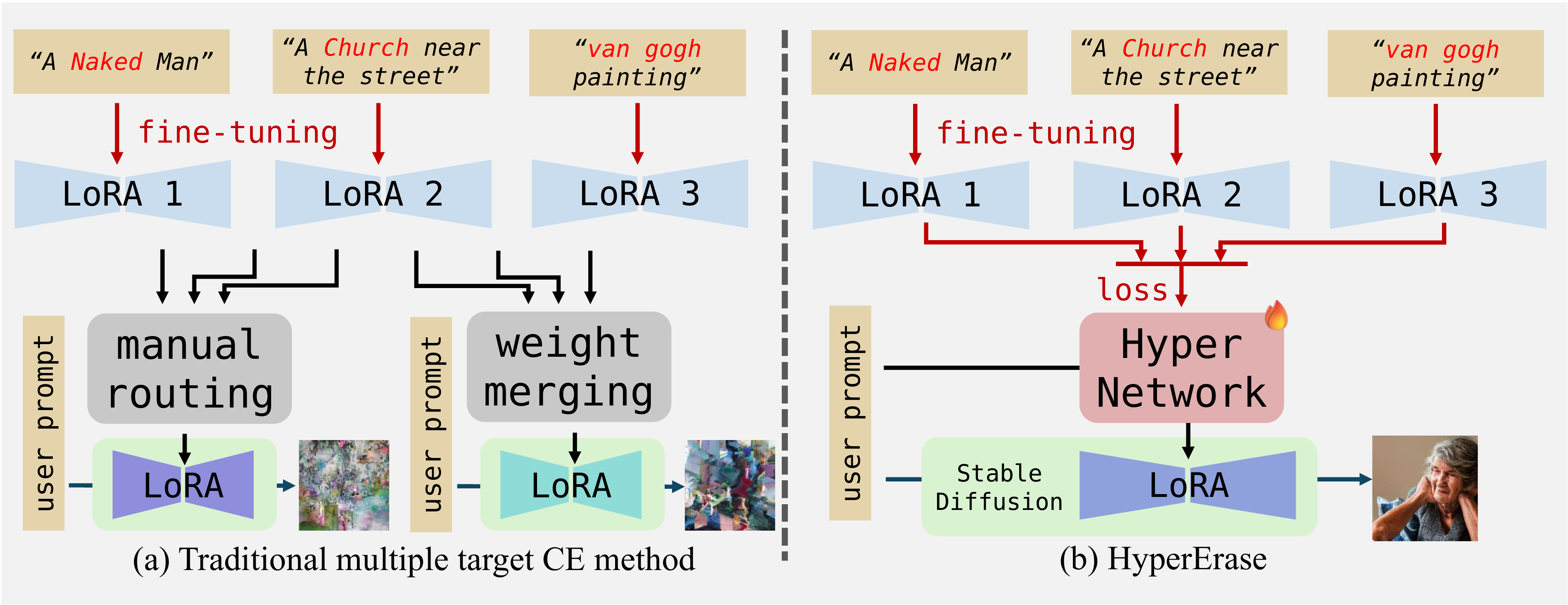}}
  \caption{Comparison of Traditional multi-concept erasure and HyperErase. HyperErase achieves high effective erasure and avoids manual routing and over-erasure by training a hypernetwork that generates dynamic LoRAs.}
  \label{fig:intro}
\end{wrapfigure}

The emergence of text-to-image (T2I) models~\cite{dhariwal2021diffusion, ho2022classifier, ho2020denoising,nichol2021glide, ramesh2022hierarchical, saharia2022photorealistic} has marked a transformative leap in visual synthesis, enabling the creation of photorealistic imagery from natural language descriptions. Yet this remarkable progress is shadowed by a critical vulnerability. The unfiltered nature of Internet-scale training corpora~\cite{milmo2023ai} frequently imbues these systems with the capacity to render objectionable or harmful visual content~\cite{jiang2023ai, roose2022ai, setty2023ai}. A straightforward remedy, excising problematic data followed by full model retraining~\cite{nichol2021glide, schramowski2023safe}, proves prohibitively costly in practice. Such an approach entails not only staggering computational demands but also the risk of unforeseen performance regressions and emergent biases~\cite{oconnor2022stable}. Concept erasure (CE) has consequently emerged as a pragmatic alternative, offering surgical and parameter-efficient interventions that selectively suppress targeted semantic concepts without compromising the model's broader generative fidelity.

Mainstream CE approaches, from early ESD~\cite{Gandikota2023ErasingCF} and UCE~\cite{gandikota2024unified}, MACE~\cite{lu2024mace} to EraseAnything~\cite{gao2025eraseanything}, follow a static weight paradigm: they output a single frozen LoRA (or fused adapter set) that bakes all designated targets into one fixed U-Net update. This works for narrow single-concept cases, but two intrinsic limits surface as erasure needs expand. First, even for a single target type, the static adapter is a forced trade-off: EraseBench~\cite{Amara2025ErasingMT} shows all such methods face a retain set dilemma, tightening erasure to block leakage degrades visually similar non-targets, while preserving fidelity lets targets slip—because one fixed weight cannot adapt to the open-ended phrasings of the same concept (e.g. "a church" vs. "a stone church on the street") without re-running gradients. When scaling to multiple concepts, this paradigm breaks further: even MACE, which trains per-concept modules and fuses them via closed-form updates to reach 100 targets, still produces a single merged static adapter. Interference accumulates between erased concepts and preserved classes as scope widens~\cite{Liu2025DyMEDM}; merging LoRAs inherently injects parameter crosstalk and direction drift~\cite{Zhang2025UnravelingLI, Li2024BlockwiseLR}, and sequential tuning causes earlier erased concepts to re-emerge~\cite{Carter2025TRACETC}. Both multi-concept erasure methods suffer from inherent bottlenecks. Storing dozens of disjoint LoRAs need manual routing and storage bloat while accepting a merged weight will under-erase targets and over-hurt non-targets.

We argue both bottlenecks stem from the core assumption that all prompts and targets must share one static weight set. We instead reframe CE as prompt-conditioned parameter amortization for a pre-defined concept family and introduce \textbf{HyperErase}. Concretely, we employ ESD as the gold baseline. For each concept in a pre-defined target family, we use ESD to erase target concept to obtain prompt-checkpoint pairs as the training dataset. We then train a single hypernetwork from this dataset. At inference, maps the input prompt to a prompt-specific LoRA for targeted erasure in one forward pass and yield near-zero updates if the prompt is outside the family . Our method removes the per-prompt gradient tax and skips LoRA merging entirely. To stabilize generation, we decouple LoRA tokens into pattern and scale subspaces, replace log-scale decoding with a square-root transform to curb multiplicative over-scaling, and apply inference-time rectification via teacher-derived canonical priors. Experiments demonstrate that our method achieves erasure performance comparable to the gold baseline in CE scenarios. Moreover, during inference, it can provide a specialized LoRA for each input prompt variation without requiring gradient updates or additional training. This enhances preservation capability compared to the gold baseline and establishes a new paradigm for CE tasks. In summary, we make the following contributions:

\begin{itemize}
    \item We recast CE as conditional parameter synthesis, where a textual concept description directly produces the erasing weight update, unifying the per-concept optimization procedures into a single amortized mapping.
    \item We propose the first hypernetwork-based CE framework and introduce a decoupled rectification strategy that stabilizes training via square-root scale encoding and restores the erasure strength of generated adapters through pattern and scale corrections.
    \item Extensive experiments across major concept categories demonstrate that our method approaches the erasure performance of gold single-concept baselines while maintaining superior generative capability.
\end{itemize}
\section{Related Works}
\label{sec:related}

Existing CE methods can be broadly categorized into dataset filtering~\cite{Rando2022RedTeamingTS,dalle3systemcard}, model retraining~\cite{schramowski2023safe}, training-based fine-tuning approaches~\cite{zhong2025closing,sun2026flowerase,sun2026floweraseopd} and training-free methods~\cite{chavhan2024conceptprune,Sun2026ActEraseAT}. Data filtering techniques (e.g., Nudenet~\cite{NotAI_NudeNet_2019} and Q16 detector~\cite{10.1145/3531146.3533192}) exclude sensitive content at the dataset level, yet retraining from scratch incurs substantial computational overhead and temporal cost. Target concepts may still permeate the latent space via ostensibly benign prompts, which in turn demands iterative retraining cycles. An alternative strategy operates directly on pre-trained models to suppress target concept generation. This paradigm has attracted considerable scholarly attention and bifurcates into training-based and training-free methodologies.

Training-based methods fine-tune pre-trained models to eliminate specific concepts. Erased Stable Diffusion (ESD)~\cite{Gandikota2023ErasingCF} aligns noise predictions between target and non-target concepts within the latent diffusion model and steers optimization through classifier-free guidance. Unified Concept Editing (UCE)~\cite{gandikota2024unified} derives closed-form solutions for cross-attention weight manipulation, modifying key and value matrices corresponding to target concept embeddings while preserving weights associated with unrelated concepts. The editing-based formulation permits near-instantaneous weight modification, enabling rapid and effective erasure. DVE~\cite{Zhang2026DifferentialVE} projects target token value vectors onto the semantic subspace spanned by non-target concepts at each denoising timestep and nullifies the residual components, enabling parameter-free concept erasure with dynamically adaptive suppression strength.

\section{Method}
\label{sec:method}

\subsection{Parameter Generator}

\begin{figure*}[t]
  \centering
  \resizebox{\linewidth}{!}{
  \includegraphics{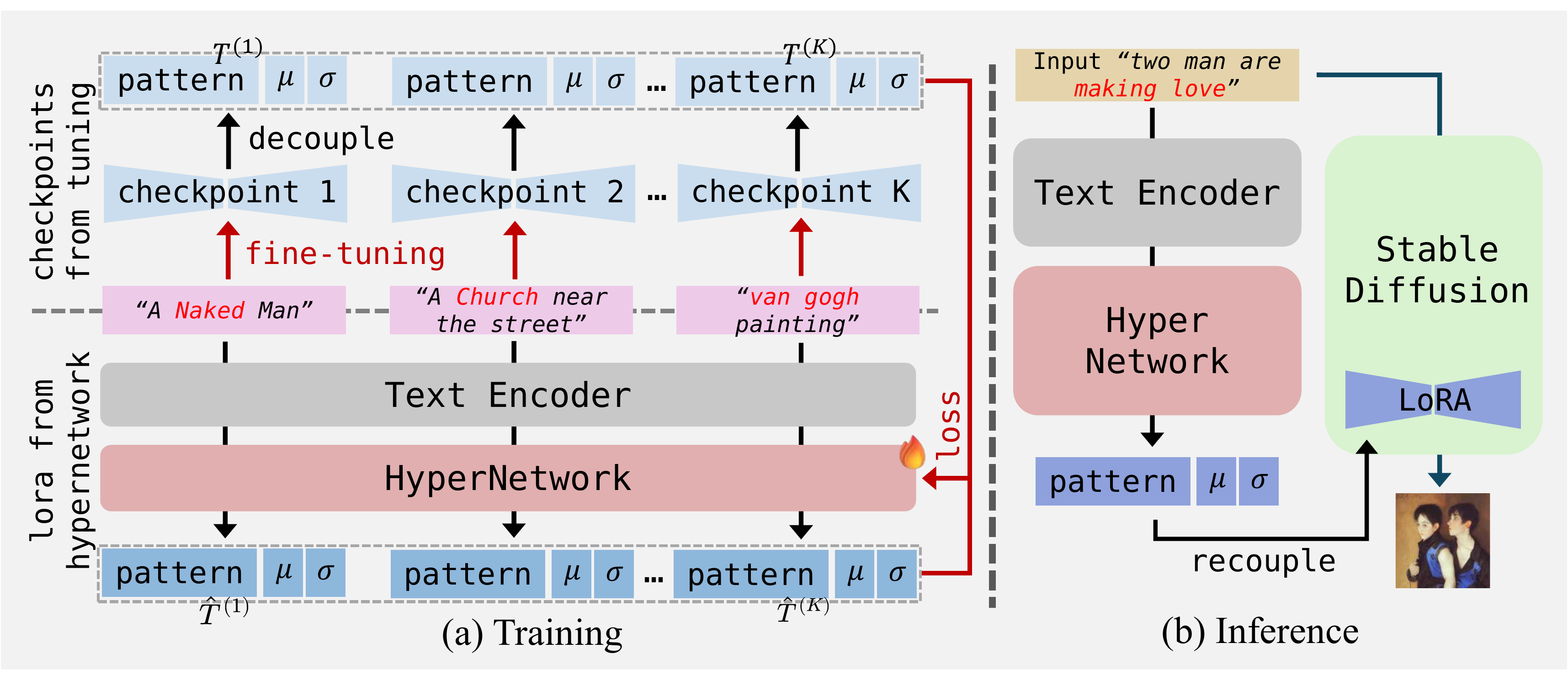}}
  \caption{Overview of HyperErase. (a) illustrates the training procedure of HyperErase. We use the gold baseline to erase a range of target concepts and, once the erasure is complete, collect checkpoint--prompt pairs; the same prompts are then fed into the HyperNetwork to fit the generation of LoRAs. To mitigate the accumulation of decay across multiple concepts, we decouple LoRA into pattern and scale dimensions and apply decoupled magnitude rectification. (b) illustrates the inference procedure. Given a new input, we use HyperErase to generate the pattern and scale checkpoints, recouple them into a LoRA, and apply it during inference.}
  \label{fig:framework}
\end{figure*}

As shown in Figure~\ref{fig:framework}, we reframe concept erasure as concept-conditional parameter amortization within a pre-defined target family. Let $c$ denote a text condition that describes a target concept. We train a parameter generator $G_\psi$ that maps $c$ to the LoRA adapter which erases that concept from a pretrained Stable Diffusion U-Net:
\begin{equation}
G_\psi(c) \;\longmapsto\; \widehat{\Theta} = \{(\hat{A}_\ell, \hat{B}_\ell)\}_{\ell=1}^{L}.
\end{equation}
Our generator is trained on adapters collected for a set of $K$ concepts, and at inference, for a new input prompt, our generator produces a corresponding LoRA. If the prompt contains the target concept, this LoRA enables precise erasure of the target concept; otherwise, it generates outputs normally.

Directly regressing raw weight matrices is impractical due to their high dimensionality and heterogeneous shapes across layers. We therefore represent all LoRA weights in a unified token grid. For each layer, the matrix $A_\ell$ (with convolutional spatial dimensions flattened) and the transposed $B_\ell^\top$ are partitioned column-wise into chunks of fixed width $t_w$. Each chunk $X \in \mathbb{R}^{r \times t_w}$ is standardized by its per-chunk mean and standard deviation:
\begin{equation}
\mu = \frac{1}{r t_w}\sum_{i,j} X_{ij}, \quad \sigma = \sqrt{\frac{1}{r t_w}\sum_{i,j} (X_{ij} - \mu)^2}, \quad \tilde{X} = \frac{X - \mu}{\sigma}.
\end{equation}
The normalized pattern $\tilde{X}$ occupies the first $r$ rows of a $h \times t_w$ token, while the scale statistics $(\mu, \sigma)$ are encoded into the remaining free rows, so that each chunk carries its directional structure and its magnitude as separable channels of a single representation. Stacking all chunks across all $L$ layers produces a target tensor $T \in \mathbb{R}^{N \times h \times t_w}$, where $N$ is determined by the U-Net architecture.

With this representation in place, training the generator reduces to supervised regression over a corpus of prompt--checkpoint pairs. For each of $K$ concepts, we collect $S$ ESD-u LoRA checkpoints from different snapshots of the ESD training trajectory and pair them with concept-specific prompt pools, so that the generator observes both the diversity of weight realizations and the diversity of natural language descriptions for every concept. A frozen BERT encoder embeds each prompt batch into a condition tensor, and a cascaded hyper-convolutional decoder expands the sequence dimension from $M$ prompt slots to $N$ weight-token slots while reshaping the spatial feature map through a sequence of separable convolutional stages, matching the token grid dimension at the terminal layer. The generator is optimized with an importance-weighted mean squared error over valid token positions:
\begin{equation}
\mathcal{L}(\psi) = \frac{1}{|\mathcal{V}|} \sum_{j \in \mathcal{V}} \omega_j \, \bigl( \widehat{T}_j - T_j \bigr)^2,
\end{equation}
where $\mathcal{V}$ denotes the set of non-padding token positions and $\omega_j$ is the normalized cross-checkpoint variance of token $j$, which assigns higher weight to positions whose values carry erasure-relevant information and down-weights near-constant padding regions.

The conventional encoding maps $\sigma$ through the logarithmic transform $e_\sigma = \log(0.9\sigma + 0.1) + b$, with the inverse recovered by exponentiation. In single-concept training this encoding is unproblematic because the scale target is identical across all samples; the network drives its prediction variance to zero and the decode becomes exact. When the generator is trained on multiple concepts simultaneously, however, the scale target varies with the concept and must be inferred from the text condition, leaving non-negligible prediction error $\varepsilon$ on the encoded value. The exponential decode $g(e) = \exp(\cdot)$ is strictly convex, so by Jensen's inequality a zero-mean error on $e_\sigma$ produces a systematic upward bias in the decoded $\sigma$:
\begin{equation}
\mathbb{E}[\hat{\sigma}] = \mathbb{E}[g(e_\sigma^* + \varepsilon)] > g(e_\sigma^*) = \sigma^*.
\end{equation}
The inflation factor is approximately $\exp(s^2/2)$ for prediction variance $s^2$, and since $A_\ell$ and $B_\ell$ are decoded independently, the individual inflations multiply through the product $\Delta W_\ell = B_\ell A_\ell$. The resulting over-scaling preserves the erasing direction but pushes the U-Net off its learned denoising manifold, corrupting generation.

We replace the logarithmic encoding with a square-root transform. Let $S$ denote a fixed constant set to the median standard deviation across all weight chunks. The scale channel is encoded as
\begin{equation}
e_\sigma = \sqrt{\sigma / S}, \qquad \hat{\sigma} = S \cdot \hat{e}_\sigma^2.
\end{equation}
The decode function is now compresses the two-decade range of LoRA standard deviations into a well-conditioned $\mathcal{O}(1)$ interval, equalizing the loss contribution across chunks whose raw magnitudes differ by orders of magnitude.

\subsection{Decoupled Magnitude Rectification}

Because each weight chunk is encoded as the pair $(\tilde{X}, (\mu, \sigma))$, where the normalized pattern $\tilde{X}$ captures the directional structure of the weight update and the per-chunk scale $(\mu, \sigma)$ determines its local magnitude. The biases of pattern and scale compound through the matrix product $\Delta W_\ell = B_\ell A_\ell$, and because the subspaces are disjoint in the token representation, they can be analyzed and corrected independently. We introduce decoupled magnitude rectification to separately rectify these two deviations.

\textbf{Pattern Renormalization: }For the pattern channels, the generator is trained with a mean squared error loss over the $S$ teacher checkpoints of each concept. These checkpoints share the same erasing objective but exhibit natural variation in their weight realizations due to the stochasticity of the ESD optimization trajectory. Under MSE regression, the optimal prediction for a given concept is the conditional expectation of the target patterns. By Jensen's inequality, the norm of the expected pattern is strictly smaller than the expected norm of the individual patterns:
\begin{equation}
\bigl\| \mathbb{E}[\tilde{X}] \bigr\|_F \;<\; \mathbb{E}\bigl[ \|\tilde{X}\|_F \bigr] = 1.
\end{equation}
This shrinkage is a property of the loss geometry and its magnitude is approximately uniform across concepts because the per-checkpoint pattern variance is comparable throughout the concept set. We address it through a renormalization that restores each predicted chunk pattern to unit Frobenius norm during detokenization:
\begin{equation}
\tilde{X}_{\text{rect}} = \frac{\tilde{X}_{\text{pred}} - \bar{\mu}_{\text{pred}}}{\bar{\sigma}_{\text{pred}}},
\end{equation}
where $\bar{\mu}_{\text{pred}}$ and $\bar{\sigma}_{\text{pred}}$ are the empirical mean and standard deviation of the predicted chunk. This operation restores the directional energy attenuated by MSE averaging without altering the orientation of the predicted pattern.

\textbf{Canonical Scale Anchoring: }For the scale channels, the generator must infer per-chunk statistics $(\mu, \sigma)$ from BERT condition vectors that simultaneously encode $K$ concepts. In this multi-concept prompt mixture, the predicted scale for any individual concept is pulled toward the cross-concept mean, attenuating concept-specific magnitude. The quadratic decode $\hat{\sigma} = S \cdot \hat{e}_\sigma^2$ further contributes a residual Jensen displacement: even with an unbiased encoded prediction, $\mathbb{E}[S \cdot \hat{e}_\sigma^2] = S \cdot (e_\sigma^{*2} + \mathrm{Var}(\hat{e}_\sigma))$, introducing a variance-dependent offset between the predicted and target standard deviations.

We address this by pre-computing concept-specific canonical scale priors and injecting them as constants into the token representation. For each concept $k$, we average the per-chunk statistics of its $S$ teacher checkpoints in raw weight space, then re-encode the averaged values through the square-root encoder:
\begin{equation}
T^{(k)}_{\text{scale}} = \mathrm{Enc}\!\left( \frac{1}{S} \sum_{i=1}^{S} \mu_i^{(k)}, \; \frac{1}{S} \sum_{i=1}^{S} \sigma_i^{(k)} \right).
\end{equation}
Averaging in raw space before encoding, rather than in the encoded space, prevents the nonlinear square-root transform from introducing a secondary bias. During training, the scale rows of the target token grid are set to the canonical block of the sampled concept:
\begin{equation}
T[:, -2:, :] \;\leftarrow\; T^{(k)}_{\text{scale}}.
\end{equation}
The generator is thereby relieved of scale regression; the per-chunk magnitudes are guaranteed to match the teacher statistics by construction, while the pattern channels continue to be learned from the full expressivity of the prompt semantics.

The canonical scale anchoring and the pattern renormalization act on disjoint subspaces of the weight token and compose additively. The former eliminates regression noise in the scale channels by replacing a learned prediction with a concept-level constant derived from the teacher distribution. The latter restores the directional norm that MSE averaging attenuates, operating purely on the normalized pattern channels. They adjust only the semantics of the representation that the tokenizer produces and consumes, making the weight-space norm of the generated LoRA adapter bias-stable with respect to the number of training concepts. This ultimately achieves efficient concept erasure for arbitrary target concepts, performing closely to the gold baseline.

\begin{figure}[htbp]
  \centering
  \includegraphics[width=\linewidth]{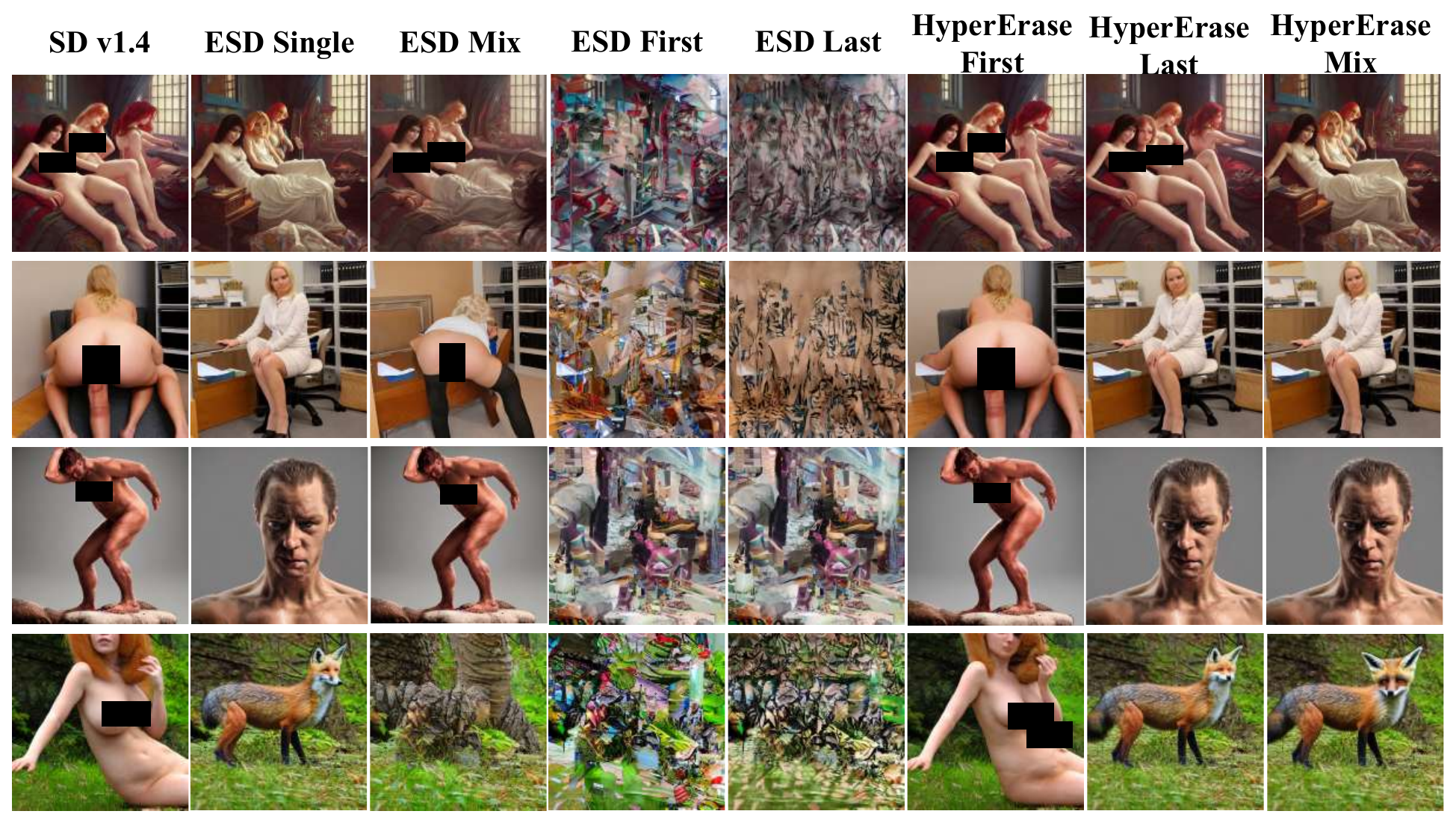}
  \caption{Visual results of \textbf{Nudity} erasure results in I2P dataset and under attacks.}
  \label{fig:exp_naked}
\end{figure}

\section{Experiments}
\label{sec:exp}

\subsection{Experimental Setup}
\label{subsec:expset}

\noindent\textbf{Baselines.} Because our method's erasure performance is related to the gold baseline, we apply our proposed method to SD v1.4 and compare it with vanilla model and the gold baseline ESD. Specifically, we compare ESD Single, ESD Mix, ESD First, ESD Last, HyperErase Mix, HyperErase First, and HyperErase Last. Here, ESD Single means using ESD to erase each concept individually and then verify. ESD Mix means using the ESD method to simultaneously erase and verify the same 14 concepts as in our method. ESD First and ESD Last sequentially erase the 14 target concepts during training. First means placing the target concept in the first position, while Last means placing the target concept in the last position for concept erasure. HyperErase First and HyperErase Last follow the same principle. This setup is mainly designed to verify the impact of sequential erasure and continual learning methods. At the same time, it eliminates unfair settings that may arise from differences in multi concept erasure methods.

\noindent\textbf{Evaluation metrics.} We evaluate on three CE tasks covering a total of 14 concepts, including nudity erasure, artist style erasure, and object erasure. For nudity erasure, we count the number of detected exposed body parts in generated images and calculate the Attack Success Rate (ASR) against adversarial attacks to measure the robustness of erasing NSFW concepts. For artist style erasure, we perform concept erasure on artist styles and calculate the Top-k classification accuracy (ACC) before and after erasure to quantify both the erasure efficacy and the preservation of non-target concepts. For object erasure, we report the ASR of the target erased concept and the ASR of non-target object concepts to measure the erasure efficacy and preservation performance. In addition, we employ CLIP Score~\cite{Radford2021LearningTV} to measure text-image consistency and FID score~\cite{Heusel2017GANsTB} to assess image quality. Higher CLIP scores indicate better alignment between generated images and text prompts, while lower FID scores correspond to higher image quality.

\noindent\textbf{Implementation details.}Following previous works~\cite{Gandikota2023ErasingCF}, we chose Stable Diffusion 1.4~\cite{rombach2022high} as the base model of all our experiments. We employ the DDIM sampler with 50 sampling steps and other hyper-parameters follow the default configurations from the respective official repositories. All experiments are performed on RTX A6000 GPUs. For more details on the specific validation dataset settings for each concept, see Appendix A.2.

\begin{table*}[htbp!]
    \centering
    \resizebox{\textwidth}{!}{%
     \begin{tabular}{cccccccccc|cc}
        \toprule
        Method  & Armputs & Belly & Buttocks & Feet & Breasts (F) & Genitalia (F) & Breasts (M) & Genitalia (M) & Total & FID ($\downarrow$) & CLIP ($\uparrow$) \\
        \midrule
            SD v1.4 & 107 & 153 & 21 & 31 & 262 & 10 & 33 & 4 & 621 & 17.84 & 31.08\\
        \midrule
            ESD Single  & 60 & 89 & 10 & 29 & 141 & 10 & 16 & 2 & 357 & 17.06 & 30.54\\
            ESD Mix  & 39 & 56 & 7 & 15 & 65 & 4 & 17 & 1 & 204 & 22.56 & 29.12\\
            ESD First & 0 & 0 & 0 & 0 & 0 & 0 & 0 & 0 & 0 & 198.93 & 17.28\\
            ESD Last & 0 & 0 & 0 & 0 & 0 & 0 & 1 & 0 & 1 & 170.79 & 18.60\\
        \rowcolor{gray!30}
            HyperErase First & 104 & 174 & 21 & 24 & 289 & 17 & 31 & 6 & 666 & 17.64 & 30.96\\
        \rowcolor{gray!30}
            HyperErase Last & 52 & 85 & 12 & 27 & 123 & 12 & 17 & 4 & 332 & 17.03 & 30.57\\
        \rowcolor{gray!30}
            HyperErase Mix & 58 & 87 & 12 & 28 & 138 & 11 & 18 & 7 & 359 & 16.96 & 30.57\\
        \bottomrule
     \end{tabular}}
    \caption{Quantity of explicit content detected using the Nudenet detector on the I2P benchmark. \textbf{F}: Female. \textbf{M}: Male. HyperErase achieves erasure performance approaching the gold baseline ESD Single while achieving better FID score and CLIP score.} 
    \label{compare-naked}
\end{table*}

\subsection{Results Analysis}
\label{subsec:exp_results}

\begin{table*}[htbp]
    \centering
    \resizebox{\linewidth}{!}{%
     \begin{tabular}{ccccccccc}
        \toprule
        Method  & I2P(\%) & MMA(\%) & Ring-16(\%) & Ring-38(\%) & Ring-77(\%) & P4D(\%) & UnDiff(\%) & Average(\%) \\
        \midrule
            ESD Single & 57.5 & 22.1 & 72.1 & 74.8 & 78.0 & 67.4 & 53.8 & 60.8\\
            ESD Mix  & 32.9 & 21.6 & 40.3 & 49.3 & 49.8 & 47.6 & 27.2 & 38.4\\
            ESD First & 0.0 & 0.0 & 0.0 & 0.0 & 0.0 & 0.0 & 0.0 & 0.0\\
            ESD Last & 0.2 & 0.0 & 0.0 & 0.0 & 0.0 & 0.0 & 0.0 & 0.03\\
         \rowcolor{gray!30}
            HyperErase First  & 107.2 & 95.6 & 107.4 & 105.3 & 114.6 & 99.1 & 102.3 & 104.5\\
        \rowcolor{gray!30}
            HyperErase Last  & 53.5 & 21.8 & 72.6 & 70.3 & 76.7 & 62.9 & 46.8 & 57.8\\
        \rowcolor{gray!30}
            HyperErase Mix  & 57.8 & 26.4 & 71.8 & 76.2 & 75.4 & 67.3 & 55.5 & 61.5\\
        \bottomrule
     \end{tabular}}
    \caption{The Attack Success Rate (ASR) against adversarial attacks in erasing NSFW concept '\textbf{Nudity}'. Our method achieves erasure performance close to the gold baseline against different types of adversarial attacks.} 
    \label{compare-attack-naked}
\end{table*}

\subsubsection{Nudity Erasure}
\label{subsubsec:nud}

Table~\ref{compare-naked} presents the experimental results of Nudity concept erasure. It can be seen that when mixing all concepts for training (HyperErase Mix), HyperErase can achieve an erasure effect close to that of the gold baseline (ESD Single). Meanwhile, we can observe that HyperErase Mix outperforms the gold baseline in both CLIP and FID scores. This is because the HyperErase method can dynamically generate personalized LoRA for newly input prompts, reducing the erasure of irrelevant concepts by static LoRA due to prompt differences. Compared to the original multi-concept training method of the gold baseline (ESD Mix), although the erasure metric of our method appears lower, the FID and CLIP Score of ESD Mix are far lower than those of the gold baseline. This indicates that the ESD Mix method directly destroys the image generation capability, thereby reducing the number of detected Exposed Bodies, rather than truly possessing better concept erasure performance. We also tested the erasure results when training by sequentially inputting concepts. Table~\ref{compare-attack-naked} shows the ASR against adversarial attacks in erasing the NSFW concept '\textbf{Nudity}'. It can be seen that our method can still achieve an erasure effect consistent with the gold baseline under multiple adversarial attacks.

\begin{wraptable}{R}{0.48\textwidth}
\vspace{-0.8em}
    \centering
    \resizebox{\linewidth}{!}{%
     \begin{tabular}{ccccc}
        \toprule
        & Real Images & ESD Single & ESD Mix & HyperErase\\
        \midrule
            LPIPS & 0.742  & 0.750 & 0.730 & 0.749  \\
        \bottomrule
     \end{tabular}}
    \caption{Comparison of LPIPS. The results of HyperErase are closer to the Real Images than other methods.}
    \label{compare_lpips}
\end{wraptable}

\begin{figure}[thbp!]
    \centering
    \begin{subfigure}[t]{0.32\textwidth}
        \centering
        \includegraphics[width=\linewidth]{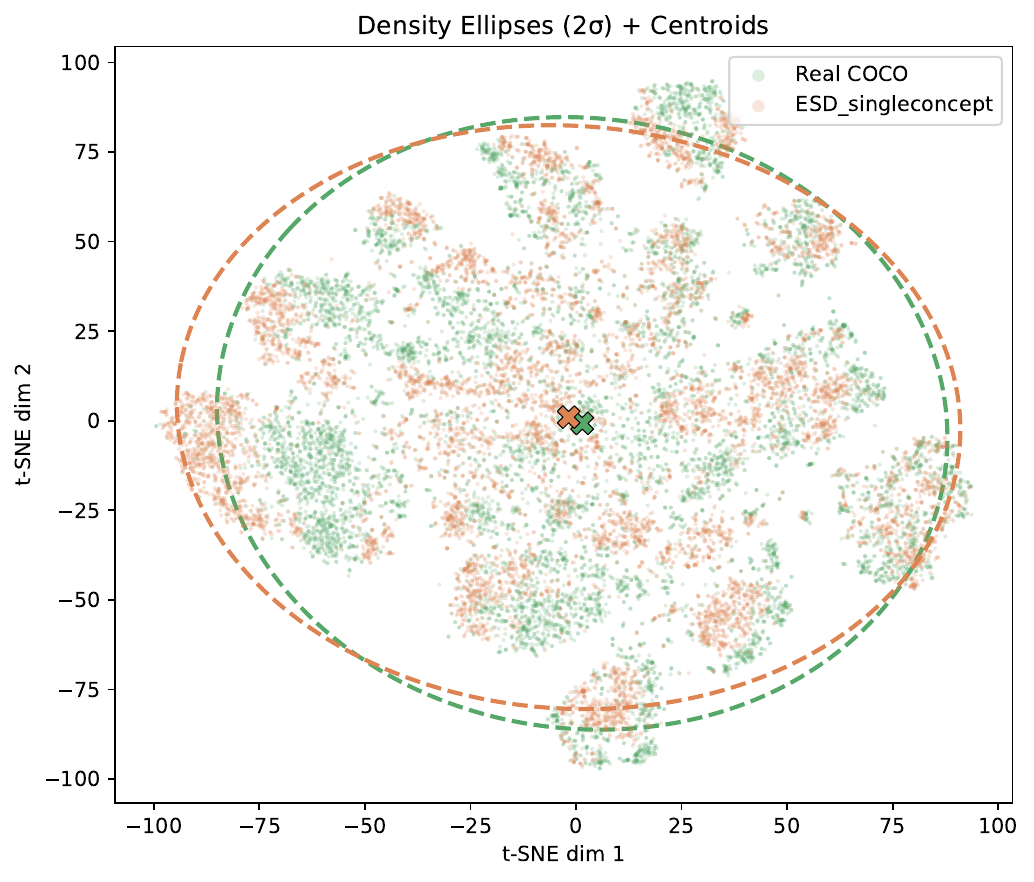}
        \caption{ESD Single}
        \label{fig:tsne_single}
    \end{subfigure}
    \hfill
    \begin{subfigure}[t]{0.32\textwidth}
        \centering
        \includegraphics[width=\linewidth]{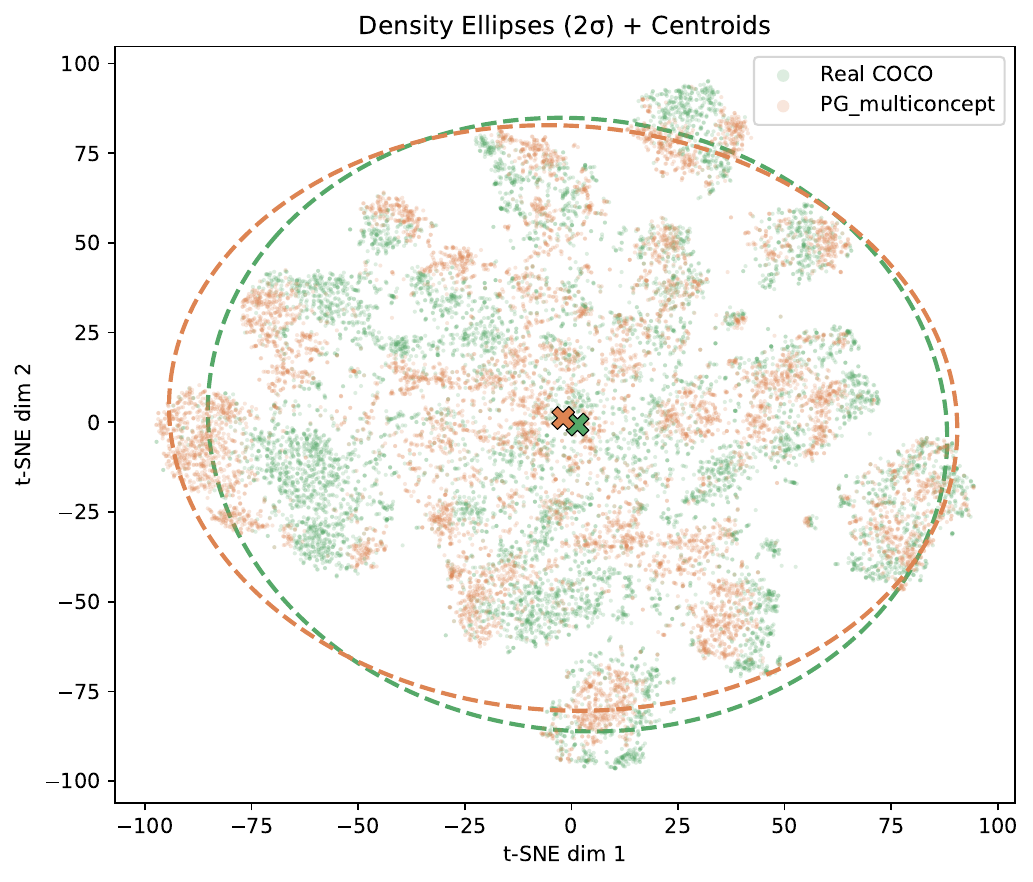}
        \caption{HyperErase}
        \label{fig:tsne-hyper}
    \end{subfigure}
    \hfill
    \begin{subfigure}[t]{0.32\textwidth}
        \centering
        \includegraphics[width=\linewidth]{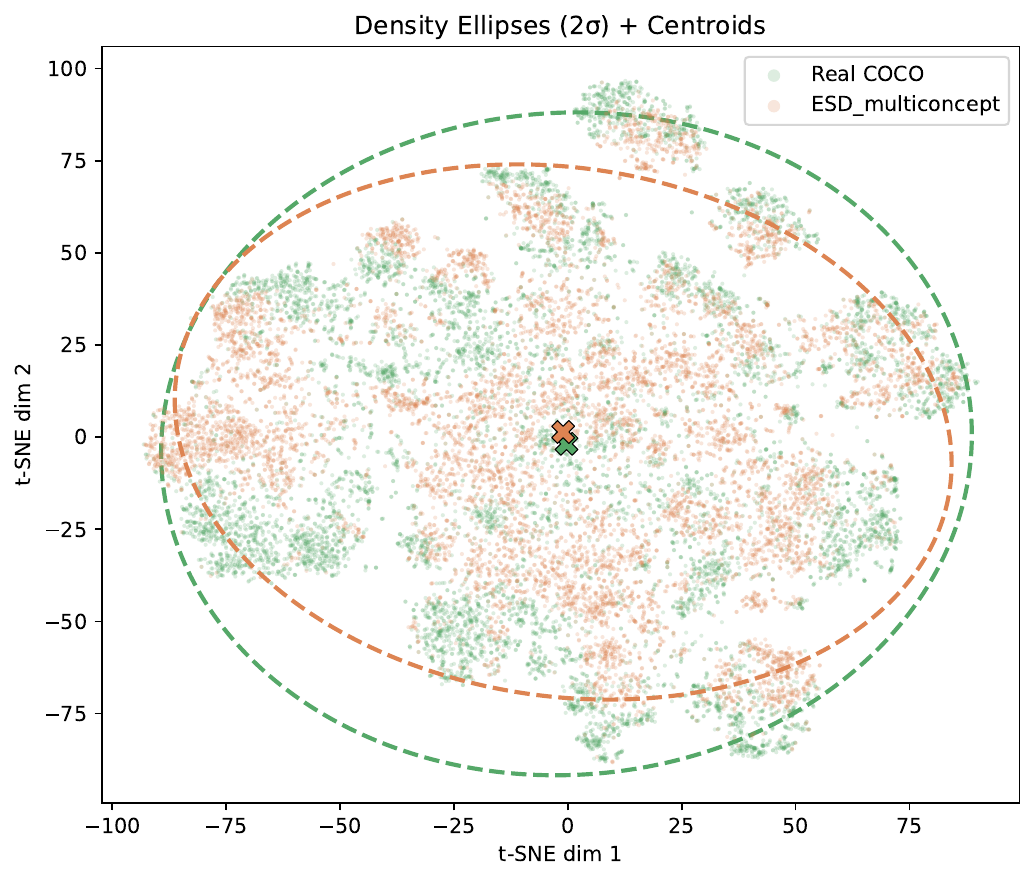}
        \caption{ESD Mix}
        \label{fig:tsne-mix}
    \end{subfigure}
    \caption{t-SNE visualization comparing generated images with real images. The t-SNE distribution of HyperErase is close to that of ESD Single, whereas ESD Mix exhibits a significant shift.}
    \label{fig_tsne}
\end{figure}

It can be seen that when using this continual learning-like method, the multi-concept method of ESD almost completely loses its generation capability, manifested as high FID and low CLIP scores. This indicates that the multi-concept method of ESD must be retrained every time a new concept is added for erasure. In contrast, the CLIP and FID computed by our method when using the continual learning-like approach show no significant difference from those of HyperErase Mix, indicating that our method does not lose generation capability when sequentially inputting new concepts for erasure, demonstrating its ability to be applied to continual learning. However, it can also be observed that when the target concept is erased at the very beginning or early stage (HyperErase First), the erasure performance degrades, even surpassing the original SD v1.4 (from 359 to 666). When the target concept is erased at the end (HyperErase Last), the erasure performance instead improves to surpass the original method. We attribute the fact that the Total Exposed Parts of HyperErase First even exceeds that of SD v1.4 to catastrophic forgetting. Because during the experiment, we directly input the trained concepts sequentially without using regularization for protection, as the gradients of the subsequent 13 concepts are continuously updated, each gradient update from subsequent tasks distorts the previously learned mappings, ultimately leading to catastrophic forgetting. In contrast, the erasure performance of HyperErase Last that surpasses the gold baseline may be because HyperErase learns meta-knowledge from erasing multiple previous concepts, and when generalizing to nudity, its effect even surpasses that of individual ESD.

To further compare the performance differences among HyperErase, ESD Mix, and ESD Single, we compute the LPIPS on the MSCOCO images generated by each method after nudity erasure and additionally render separate t-SNE plots for the respective outputs. As shown in Table~\ref{compare_lpips} and Figure~\ref{fig_tsne}, ESD Mix yields LPIPS values lower than those of the real images, which likely stems from mode collapse or a notable reduction in diversity, and its t-SNE embedding also exhibits a marked deviation. By contrast, HyperErase's LPIPS and t-SNE distributions both align substantially closer to the gold baseline, and its LPIPS values are in fact even nearer to the real images than the gold baseline itself, demonstrating that HyperErase genuinely enhances generative capability rather than merely overfitting.

\subsubsection{Object Erasure}
\label{subsubsec:obj}

\begin{table}[ht]
    \centering
    \resizebox{0.6\linewidth}{!}{%
     \begin{tabular}{ccccc}
        \toprule
        Method  & ASR\textsubscript{e}(\%) & ASR\textsubscript{k}(\%) & FID ($\downarrow$) & CLIP ($\uparrow$) \\
        \midrule
            ESD Single  & 3.60 & 62.54 & 18.41 & 30.29\\
            ESD Mix  & 18.10 & / & 22.56 & 29.12\\
            ESD First  & 0.00 & / & 183.84 & 17.59\\
            ESD Last  & 0.00 & / & 198.74 & 17.14\\
            \rowcolor{gray!30}
            HyperErase First  & 69.15 & 70.56 & 18.60 & 30.28\\
            \rowcolor{gray!30}
            HyperErase Last  & 3.98 & 62.49 & 18.38 & 30.37\\
            \rowcolor{gray!30}
            HyperErase Mix  & 3.92 & 62.84 & 18.24 & 30.35\\
        \bottomrule
     \end{tabular}}
    \caption{Comparison of ASR\textsubscript{e}, ASR\textsubscript{k}, FID and CLIP Score for Object concepts erasure tasks.} 
    \label{compare_object}
\end{table}

Table~\ref{compare_object} compares the average erasure performance of different methods on the Object Erasure task. Since ESD Mix and ESD Last/First simultaneously erase all Object concepts and generate a static LoRA, there are no Non-target Concepts. To evaluate our method, we use HyperErase with prompts containing target concepts to generate personalized LoRA, while simultaneously using this LoRA and Non-target prompts to generate images and calculate ASR\textsubscript{k}, thereby evaluating the erasure performance of our method. Similar to Nudity Erasure, HyperErase can achieve an erasure effect close to that of ESD Single and possesses better CLIP and FID. It is worth noting that the personalized LoRA generated using prompts with target concepts still maintains the same non-target concept generation capability as ESD Single. This indicates that our method can not only reduce interference with irrelevant concepts by dynamically generating LoRA, but the generated LoRA itself also introduces less influence to irrelevant concepts.

\begin{figure}[htbp]
  \centering
  \begin{minipage}[h]{0.48\textwidth}
    \centering
    \includegraphics[width=\linewidth]{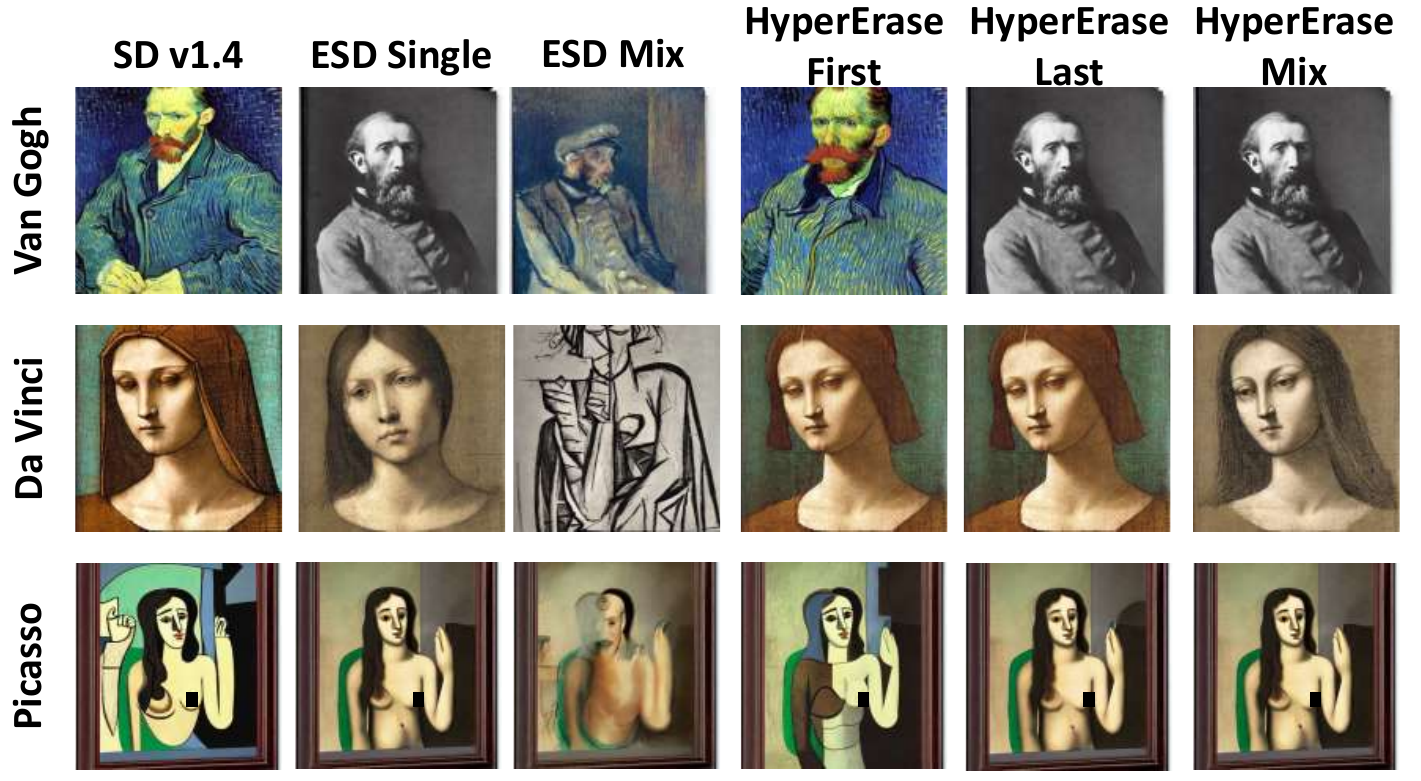}
    \caption{Visual results of artist style erasure results.}
    \label{fig:exp_style}
  \end{minipage}
  \hfill
  \begin{minipage}[h]{0.48\textwidth}
    \centering
    \includegraphics[width=\linewidth]{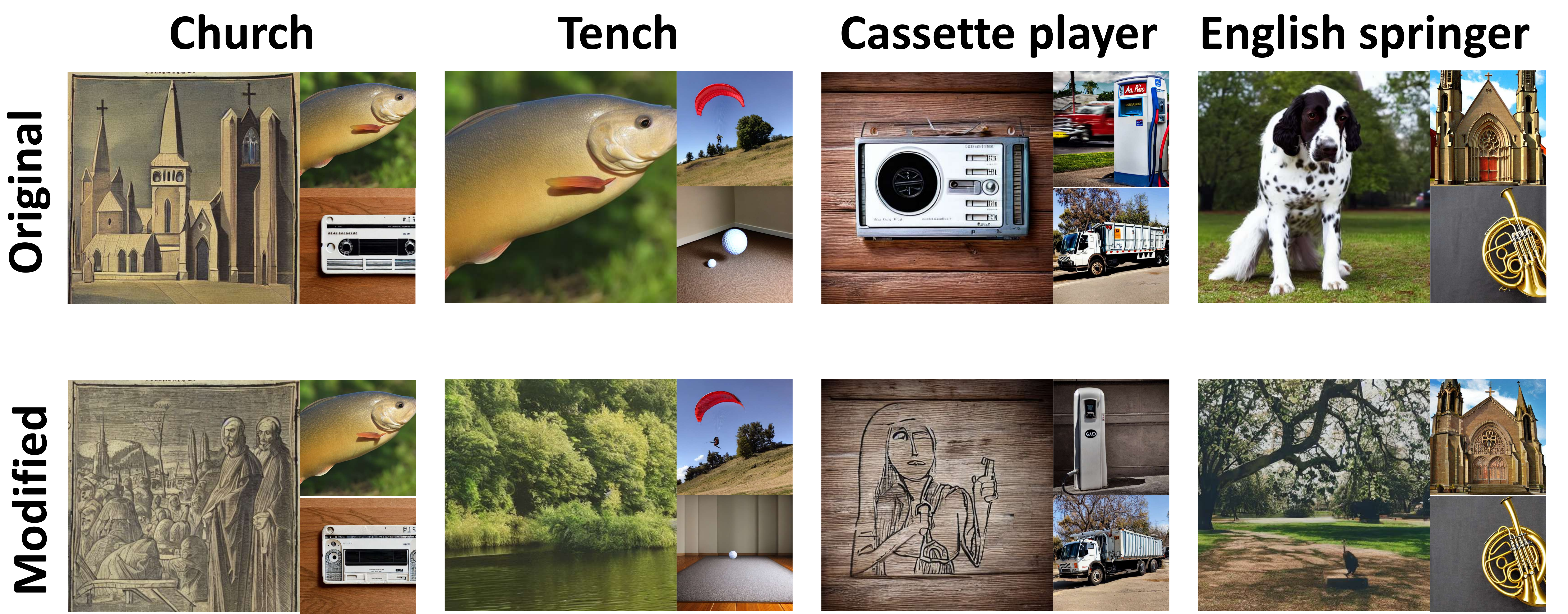}
    \caption{Visual results of object concepts erasure.}
    \label{fig:exp_object}
  \end{minipage}
\end{figure}

\subsubsection{Artist Style Erasure}
\label{subsubsec:style}

The artist style erasure results are shown in Table~\ref{compare-style}. Similar to Object Erasure and Nudity Erasure, our method exhibits an erasure effect close to that of ESD Single, demonstrating that our method remains effective in the Artist Style Erasure task.

\begin{table*}[htbp]
    \centering
    \resizebox{\textwidth}{!}{%
     \begin{tabular}{lcccccc|cccccc|cccccc}
        \toprule
        & \multicolumn{6}{c|}{Erase \textbf{Van Gogh}} & \multicolumn{6}{c|}{Erase \textbf{Leonardo Da Vinci}} & \multicolumn{6}{c}{Erase \textbf{Pablo Picasso}}\\
        \cmidrule(lr){2-7} \cmidrule(lr){8-13} \cmidrule(lr){14-19}
        \multirow{2}{*}{Method} & \multicolumn{3}{c}{ACC} & \multirow{2}{*}{FID ($\downarrow$)} & \multirow{2}{*}{CLIP ($\uparrow$)} & & \multicolumn{3}{c}{ACC} & \multirow{2}{*}{FID ($\downarrow$)} & \multirow{2}{*}{CLIP ($\uparrow$)} & & \multicolumn{3}{c}{ACC} & \multirow{2}{*}{FID ($\downarrow$)} & \multirow{2}{*}{CLIP ($\uparrow$)} \\
        \cmidrule(lr){2-4} \cmidrule(lr){8-10} \cmidrule(lr){14-16}
        & Top-1 & Top-3 & Top-5 & & & & Top-1 & Top-3 & Top-5 & & & & Top-1 & Top-3 & Top-5 & & \\
        \midrule
        SD v1.4 & 0.74 & 0.94 & 1.00 & 17.84 & 31.08 & & 0.00 & 0.10 & 0.26 & 17.00 & 31.38 & & 0.2 & 0.98 & 1 & 17.00 & 31.38 &\\
        \midrule
        ESD Single & 0.22 & 0.48 & 0.52 & 17.46 & 30.92 & & 0.00 & 0.02 &  0.10 & 17.56 & 30.89 & & 0.04 & 0.54 & 0.76 & 17.65 & 30.84 &\\
        ESD Mix & 0.26 & 0.56 & 0.70 & 22.56 & 29.12 & & 0.00 & 0.04 & 0.10 & 22.56 & 29.12 & & 0.08 & 0.70 & 0.70 & 22.56 & 29.12 &\\
        ESD First & 0.00 & 0.00 & 0.02 & 117.74 & 20.28 & & 0.00 & 0.00 & 0.00 & 183.32 & 18.28 & & 0.00 & 0.04 & 0.10 & 182.80 & 17.67 &\\
        ESD Last & 0.00 & 0.00 & 0.02 & 155.80 & 17.77 & & 0.00 & 0.00 & 0.00 & 176.88 & 17.63 & & 0.00 & 0.02 & 0.14 & 186.44 & 18.46 &\\
        \rowcolor{gray!30}
        HyperErase First & 0.84 & 0.94 & 0.98 & 16.61 & 30.54 & & 0.00 & 0.14 & 0.20 & 16.75 & 30.50 & & 0.48 & 0.98 & 0.98 & 16.79 & 30.52 &\\
        \rowcolor{gray!30}
        HyperErase Last & 0.22 & 0.52 & 0.60 & 17.47 & 30.93 & & 0.00 & 0.00 & 0.10 & 17.41 & 30.90 & & 0.08 & 0.58 & 0.78 & 17.05 & 30.81 &\\
        \rowcolor{gray!30}
        HyperErase Mix & 0.16 & 0.44 & 0.48 & 17.57 & 30.90 & & 0.00 & 0.04 & 0.10 & 16.91 & 30.68 & & 0.08 & 0.70 & 0.76 & 17.36 & 30.84 &\\ 
        \bottomrule
     \end{tabular}}
     \caption{Comparison with artist concepts erasure results. ACC represents the top-k classification accuracy. } 
    \vspace{-0.8em}
    \label{compare-style}
\end{table*}

\subsection{Further Analysis}
\label{subsec:further}

\noindent\textbf{Ablation on Different Base Modules: }We apply HyperErase to the Flow Matching Model-based FLUX.1-dev model~\cite{flux2024}. We first use ESD-u to erase the "Nudity" concept and train it with a low learning rate after the erasure is completed to collect LoRA-prompt pairs. Subsequently, we train HyperErase using the obtained dataset. For FLUX.1-dev, ESD, and HyperErase, we use the I2P dataset and 10,000 prompts from MSCOCO to generate images, respectively, and calculate the total exposed body, CLIP, and FID scores. As shown in Table~\ref{compare-model}, although the erasure effect shows a certain degree of decline, HyperErase can still successfully erase the concept on FLUX.1-dev, and the FID and CLIP are consistent with the results of the SD v1.4 model.

\begin{table}[htbp]
    \centering
    \begin{minipage}[t]{0.48\linewidth}
        \centering
        \resizebox{1.0\linewidth}{!}{%
        \begin{tabular}{c|ccc}
            \toprule
             Model & Total($\downarrow$) & FID ($\downarrow$) & CLIP ($\uparrow$) \\
            \midrule
                FLUX 1.dev & 710 & 26.63 & 30.88\\
                ESD Single & 286 & 24.11 & 30.83\\
            \rowcolor{gray!30}
                HyperErase  & 316 & 24.33 &  30.84\\
            \bottomrule
        \end{tabular}}
        \caption{Ablation results on FLUX.1-dev.} 
        \label{compare-model}
    \end{minipage}
    \hfill
    \begin{minipage}[t]{0.48\linewidth}
        \centering
        \resizebox{1.0\linewidth}{!}{%
        \begin{tabular}{cccc}
            \toprule
                Type & Total & FID ($\downarrow$) & CLIP ($\uparrow$) \\
            \midrule
                SD v1.4 & 621 & 17.84 & 31.08\\
                \midrule
                \textit{ESD Single}  & 357 & 17.06 & 30.54 \\
                \textit{without decouple} & 464 & 17.77 & 30.35\\
                \rowcolor{gray!30}
                \textit{with decouple} & 359 & 16.96 & 30.57\\
                \bottomrule
        \end{tabular}}
        \caption{Ablation results on HyperErase without and with decouple.} 
        \label{compare-dmc}
    \end{minipage}
\end{table}

\noindent\textbf{Ablation on Decoupled Magnitude Rectification: }To demonstrate the effectiveness of our proposed decouple magnitude rectification, we trained 14 concepts using the same training dataset without applying decouple magnitude rectification, and validated the erasure effect and generation quality using I2P and MSCOCO. The results are shown in Table~\ref{compare-dmc}. It can be observed that without applying decouple magnitude rectification, the decay of multiple concepts accumulates, ultimately leading to a significant degradation in erasure capability. In contrast, the erasure effect after applying decouple magnitude rectification is close to that of the gold baseline.

\section{Concluasion}

In this work, we presented HyperErase, a  hypernetwork-driven framework for multi-concept erasure in T2I models. By reformulating concept erasure as prompt-conditioned parameter amortization and combining  decoupled magnitude rectification, Hyper-Erase effectively synthesizes prompt-specific adapters that eliminate target concepts while preserving the model’s overall generative fidelity and semantic alignment. Extensive experiments across diverse erasure settings, including nudity, artist style, and object concepts, demonstrate that our method consistently matches the gold single-concept baseline in erasure effectiveness while improving the trade-off with image quality and semantic alignment. Quantitative evaluations and qualitative analyses further show that the resulting models maintain strong retention of non-target concepts and exhibit robust resistance to various adversarial attacks. These results highlight that HyperErase establishes a new and promising paradigm for concept erasure.

\bibliography{iclr2027_conference}
\bibliographystyle{iclr2027_conference}

\clearpage
\appendix

\section{Details of Implementation}
\label{app:imp}

\subsection{Details of Hyper-parameters}
\label{appendix:hyper-param}

For all 14 concept-erasure tasks, the following settings remain identical. We use Stable Diffusion v1.4 as the pretrained backbone and train the parameter generator $G_\psi$ to map a text condition $c$ to a LoRA adapter for the U-Net. For each concept, we retain 50 ESD-u teacher checkpoints and sample $M = 32$ text prompts per
training step. The prompts are tokenized to a maximum length of 128 and encoded with a frozen all-MiniLM-L12-v2 Sentence-BERT encoder, whose hidden dimension is 384. During training, each LoRA checkpoint is represented by a
token tensor $T \in \mathbb{R}^{N \times h \times t_w}$, where $N = 7{,}502$, $h = 10$, and $t_w = 130$. The normalized pattern is denoted by $\tilde{X}$ and the per-chunk scale statistics by $(\mu, \sigma)$. Canonical scale anchoring is enabled for each concept $k$, the scale rows are replaced by the raw-space mean of its 50 teacher checkpoints and then re-encoded with the square-root transform. Thus the training target follows

\begin{equation}
T_{\mathrm{scale}}^{(k)} = \mathrm{Enc}(\frac{1}{S}\sum_{i=1}^{S} \mu_i^{(k)},
\frac{1}{S}\sum_{i=1}^{S} \sigma_i^{(k)}), \quad S = 50
\end{equation}

and the scale rows of $T$ are set to $T_{\mathrm{scale}}^{(k)}$. The generator
is optimized with the importance-weighted mean-squared objective $\mathcal{L}(\psi)$ defined in the paper. The criterion weights are computed with the square-root encoding and clamped to a floor of 0.05 before renormalization. Training uses fp32 data, the AdamW optimizer, a learning rate of $3 \times 10^{-5}$. The implementation uses standard AdamW by default. During inference, we use the DDIM sampler with 50 denoising steps and a classifier-free guidance scale of 7.5. The details of Hyper-parameters are listed in Table~\ref{tab:all14-sqrt-decouple-hparams}

\begin{table*}[htbp!]
    \centering
    \small
    \begin{tabular}{cc}
        \hline
        \textbf{Hyper-parameters} & \textbf{Value} \\
        \hline
        $K$ (number of concepts) & $14$ \\
        $S$ (teacher checkpoints per concept) & $50$ \\
        $M$ (text prompts per training step) & $32$ \\
        Maximum text length & $128$ \\
        Condition dimension & $384$ \\
        $N$ (number of weight tokens) & $7{,}502$ \\
        $h$ (token height) & $10$ \\
        $t_w$ (token width) & $130$ \\
        $S$ (sqrt-encoding normalizer) & $0.006$ \\
        Learning rate & $3 \times 10^{-5}$ \\
        Noise enhancement coefficient & $1 \times 10^{-4}$ \\
        Sampling steps & $50$ \\
        Classifier-free guidance scale & $7.5$ \\
        \hline
    \end{tabular}
    \caption{Details of Hyper-parameter values.}
    \label{tab:all14-sqrt-decouple-hparams}
\end{table*}

\subsection{Details of Validation Dataset}

We unify the detailed experimental settings for erasing and validating each concept as follows:

\textbf{Nudity Erasure: }We erase the Nudity concept and generate images with all 4703 prompts and evaluation seeds from the I2P dataset~\cite{schramowski2023safe}. To evaluate robustness against prevailing adversarial attacks, we further conduct experiments on MMA-diffusion\cite{Yang2023MMADiffusionMA}, Prompt4Debugging (P4D)\cite{Chin2023Prompting4DebuggingRT}, Ring-a-bell\cite{Tsai2023RingABellHR} and UnlearnDiff\cite{Zhang2023ToGO}. We measure the attack success rate (ASR) and robustness by employing the Nudenet detector~\cite{NotAI_NudeNet_2019} to count the number of exposed body parts in images generated before and after erasure. 

\textbf{Object Erasure: }We erase 10 object concepts from ImageNet~\cite{2009ImageNet} to assess how effectively target object concepts are removed. We generate 500 images for each object with the same prompt template (i.e. "an image of a <object>") and detect target object by a ResNet-50 ImageNet classifier~\cite{He2015DeepRL}. We then calculate the Attack Success Rate (ASR) for the target concept as well as for the other nine non-target concepts.

\textbf{Artist Style Erasure: }We assess the performance of style erasure on three artist styles, Van Gogh, Da Vinci, and Picasso and validate and evaluate using the validation dataset provided by Conceptprune~\cite{chavhan2024conceptprune}. We apply the style classifier provided by UnlearnDiff to categorize the images produced, and report the Top-k accuracy to measure how well the style is removed. 

To evaluate whether the model retains its general usefulness, we also generate 10,000 images with prompts drawn from MS-COCO and then calculate both CLIP and FID scores for every method.

\begin{table}[htpb]
    \centering
    \resizebox{\linewidth}{!}{%
     \begin{tabular}{c|cccc|cccc}
        \toprule
        Method & \multicolumn{4}{c}{ESD Single} & \multicolumn{4}{|c}{HyperErase Mix} \\
        \cmidrule(lr){2-5} \cmidrule(lr){6-9}
        Concept & ASR\textsubscript{e}(\%) & ASR\textsubscript{k}(\%) & FID ($\downarrow$) & CLIP ($\uparrow$) & ASR\textsubscript{e}(\%) & ASR\textsubscript{k}(\%) & FID ($\downarrow$) & CLIP ($\uparrow$)  \\
        \midrule
            Vanilla & / & / & 17.84 & 31.08 & / & / & 17.84 & 31.08 \\
        \midrule
            Church  & 15.0 & 71.8 & 17.64 & 30.69 & 19.0 & 72.0 & 17.56 & 30.72 \\
            Tench  & 0.0 & 56.6 & 17.70 & 30.32 & 0.0 & 57.2 & 17.75 & 30.35 \\
            Golf Ball  & 0.0 & 57.4 & 17.10 & 30.50 & 0.0 & 57.2 & 16.93 & 30.51 \\
            English Springer  & 0.2 & 65.9 & 18.61 & 30.28 & 0.2 & 66.0 & 18.40 & 30.32 \\
            Cassette Player  & 0.0 & 66.9 & 17.43 & 30.16 & 0.0 & 68.9 & 17.00 & 30.26 \\
            Chain Saw  & 0.8 & 56.2 & 19.18 & 29.95 & 0.6 & 56.4 & 19.08 & 29.98 \\
            French Horn  & 0.2 & 58.3 & 17.71 & 30.47 & 0.2 & 58..8 & 17.57 & 30.52 \\
            Garbage Truck & 16.0 & 63.2 & 19.93 & 30.19 & 15.6 & 62.7 & 19.68 & 30.32 \\
            Gas Pump  & 1.4 & 65.3 & 20.53 & 30.07 & 1.4 & 65.4 & 20.18 & 30.13 \\
            Parachute & 2.4 & 63.7 & 18.30 & 30.31 & 2.2 & 63.7 & 18.21 & 30.35 \\
        \bottomrule
     \end{tabular}}
    \caption{Details of each object erasure results. We report 10 object and list the Top-1 ASR. \textbf{ASR\textsubscript{e}} represents the ASR of target concept that should be erased, while \textbf{ASR\textsubscript{k}} represents the ASR of other concepts that need to be kept. Compared with the gold baseline ESD Single, HyperErase achieves comparable erasure effectiveness for every concept while reducing the impact on unrelated concepts, as reflected by improved CLIP score and a lower FID.} 
    \label{compare_target_object}
\end{table}

\section{Additional Experiment Results}
\label{app:exp}

\subsection{Details of Object Erasure}

To better demonstrate the effectiveness of HyperErase, we compare the detailed erasure results for each concept between HyperErase Mix and ESD Single in Table~\ref{compare_target_object}. It can be seen that our method outperforms the gold baseline in both CLIP and FID scores across all concepts. The ASR\textsubscript{e} and ASR\textsubscript{k} are either close to or partially better than the gold baseline, indicating that our method can effectively learn from the gold baseline to achieve multi-concept erasure.

\subsection{Additional Ablation Study}

\noindent\textbf{Ablation on The Epoch: } We evaluated the erasure performance of HyperErase under different epochs. As shown in Table~\ref{compare_number_of_epoch}, as the number of epochs increases, the exposed body decreases, while FID and CLIP remain unchanged, indicating that the erasure performance improves with the increase of epochs. It is worth noting that the erasure performance at 2000 epochs is already quite close to the gold baseline, indicating that HyperErase can achieve a certain effect without training for a very large number of epochs.

\noindent\textbf{Ablation on The Number of Concepts: }Under the SD v1.4 framework, we employ HyperErase to simultaneously erase different numbers of concepts to verify the impact of the number of concepts on the erasure effect. Specifically, we perform concept erasure on 3, 5, 10, and all 14 concepts, calculate the exposed body for Nudity, and use 10,000 prompts randomly sampled from the MSCOCO dataset to evaluate the FID and CLIP scores, so as to measure the overall image quality and image-text alignment. As shown in Table~\ref{compare_number_of_multi}, in the multi-concept erasure scenario, as the number of concepts increases, the exposed body only decreases slightly and still remains close to the baseline. Meanwhile, under different numbers of erased concepts, the CLIP and FID scores are basically consistent. It is worth noting that when the number of concepts is small, the exposed body is even less than that of the gold baseline. We speculate that this is caused by overfitting. To control variables, the HyperErase with different numbers of concepts are all trained for the same number of epochs, so overfitting may occur when the number of concepts is small. Moreover, the erasure effect itself improves with the increase of epochs, as shown in the Tabel~\ref{compare_number_of_epoch}.

\begin{table}[ht]
    \centering
    \begin{minipage}[t]{0.48\linewidth}
        \centering
        \resizebox{\linewidth}{!}{%
        \begin{tabular}{cccc}
            \toprule
            Method & Total & FID ($\downarrow$) & CLIP ($\uparrow$) \\
            \midrule
            SD v1.4                          & 621 & 17.84 & 31.08\\
            \midrule
            ESD Single                       & 357 & 17.06 & 30.54\\
            HyperErase \textit{3 concepts}   & 348 & 17.04 & 30.57\\
            HyperErase \textit{5 concepts}   & 349 & 17.02 & 30.59\\
            HyperErase \textit{10 concepts}  & 355 & 16.96 & 30.58\\
            HyperErase \textit{14 concepts}  & 359 & 16.69 & 30.57\\
            \bottomrule
        \end{tabular}}
        \captionof{table}{Comparison of total exposed body, FID and CLIP Score for multiple erasure tasks with different numbers of concepts.}
        \label{compare_number_of_multi}
    \end{minipage}
    \hfill%
    \begin{minipage}[t]{0.48\linewidth}
        \centering
        \resizebox{\linewidth}{!}{%
        \begin{tabular}{cccc}
            \toprule
            Method & Total & FID ($\downarrow$) & CLIP ($\uparrow$) \\
            \midrule
            SD v1.4                          & 621 & 17.84 & 31.08\\
            \midrule
            ESD Single                       & 357 & 17.06 & 30.54\\
            HyperErase w/ 2000 epoch         & 361 & 16.96 & 30.57\\
            HyperErase w/ 4000 epoch         & 361 & 16.99 & 30.57\\
            HyperErase w/ 6000 epoch         & 357 & 16.96 & 30.57\\
            HyperErase w/ 8000 epoch         & 359 & 16.69 & 30.57\\
            \bottomrule
        \end{tabular}}
        \captionof{table}{Comparison of total exposed body, FID and CLIP Score for different training epochs.}
        \label{compare_number_of_epoch}
    \end{minipage}
\end{table}

\noindent\textbf{Ablation on The Gold Baseline and Modules: }Our method is agnostic to both the gold baseline and the modules of the model. To verify this claim, we adopt another concept erasure method, EraseAnything, as the gold baseline. We utilize EraseAnything to fine-tune the model to obtain the erased LoRA, and sample multiple erased parameters along with their corresponding prompts as training data to train our HyperErase. Subsequently, we use all prompts from the I2P dataset and 10k prompts from MSCOCO to generate images, and evaluate them by detecting exposed body parts and measuring the CLIP score and FID. In the main experiments, we used ESD-U as the gold baseline, so our method consistently generates LoRA parameters for all non-cross-attention (non-CA) layers. In contrast, EraseAnything performs fine-tuning on the text-projection layers; therefore, we generate LoRA parameters for the corresponding layers. The experimental results are shown in Table~\ref{compare_gold_baseline}. It can be seen that our method can still achieve concept erasure performance close to the gold baseline, which demonstrates that our method does not depend on a specific gold baseline method and is independent of the model's modules.

\begin{table}[htbp]
    \centering
    \begin{minipage}[t]{0.48\textwidth}
        \centering
        \resizebox{\linewidth}{!}{%
        \begin{tabular}{cccc}
            \toprule
            Methods & Total & FID ($\downarrow$) & CLIP ($\uparrow$) \\
            \midrule
            SD v1.4        & 621   & 17.84 & 31.08 \\
            EraseAnything  & 315   & 18.26 & 30.69 \\
            HyperErase     & 337   & 18.28 & 30.70 \\
            \bottomrule
        \end{tabular}}
        \caption{Comparison of total export parts, FID and CLIP Score for Baseline, EraseAnything and HyperErase. Our method can still generate erasure LoRA with different gold baseline and modules.}
        \label{compare_gold_baseline}
    \end{minipage}
    \hfill
    \begin{minipage}[t]{0.48\textwidth}
        \centering
        \resizebox{\linewidth}{!}{%
        \begin{tabular}{cccc}
            \toprule
            Methods & Total & FID ($\downarrow$) & CLIP ($\uparrow$) \\
            \midrule
            SD v1.4        & 621   & 17.84 & 31.08 \\
            RECE           & 68    & 17.74 & 30.66 \\
            HyperErase(Best)     & 73   & 17.48 & 30.88 \\
            \bottomrule
        \end{tabular}}
        \caption{Comparison of total export parts, FID and CLIP Score for Baseline, RECE and HyperErase. The best results show that HyperErase can still generate erasure LoRA with non-LoRA gold baseline.}
        \label{compare_rece}
    \end{minipage}
\end{table}

\noindent\textbf{Ablation on non-LoRA Gold Baseline: } In previous experiments, the gold baseline we used was to train a LoRA for erasure. To verify the applicability of our method, we adopt a training-free method, RECE, as the gold baseline. RECE~\cite{Gong2024ReliableAE} directly derives a new embedding for the target concept in the cross-attention layers based on a closed-form solution and aligns it to a harmless concept. Formally, RECE directly edits the weights of the corresponding layers without generating a LoRA. Therefore, we first perform erasure using RECE and obtain the erased weights. Subsequently, we use these weights and the original weights to approximate a LoRA, and then use this LoRA to train HyperErase. The results are shown in Table~\ref{compare_rece}. It can been seen that HyperErase can still generate erased LoRA. However, we also observe that fitting RECE is not always successful. In several experiments, there are cases where the exposed parts reach 301, which is much higher than that of RECE. We speculate that there is a problem with the weight-to-LoRA conversion method, which may require targeted tuning.

\subsection{Additional Visual Results}

\begin{figure*}[htbp]
  \centering
  \resizebox{\linewidth}{!}{
  \includegraphics{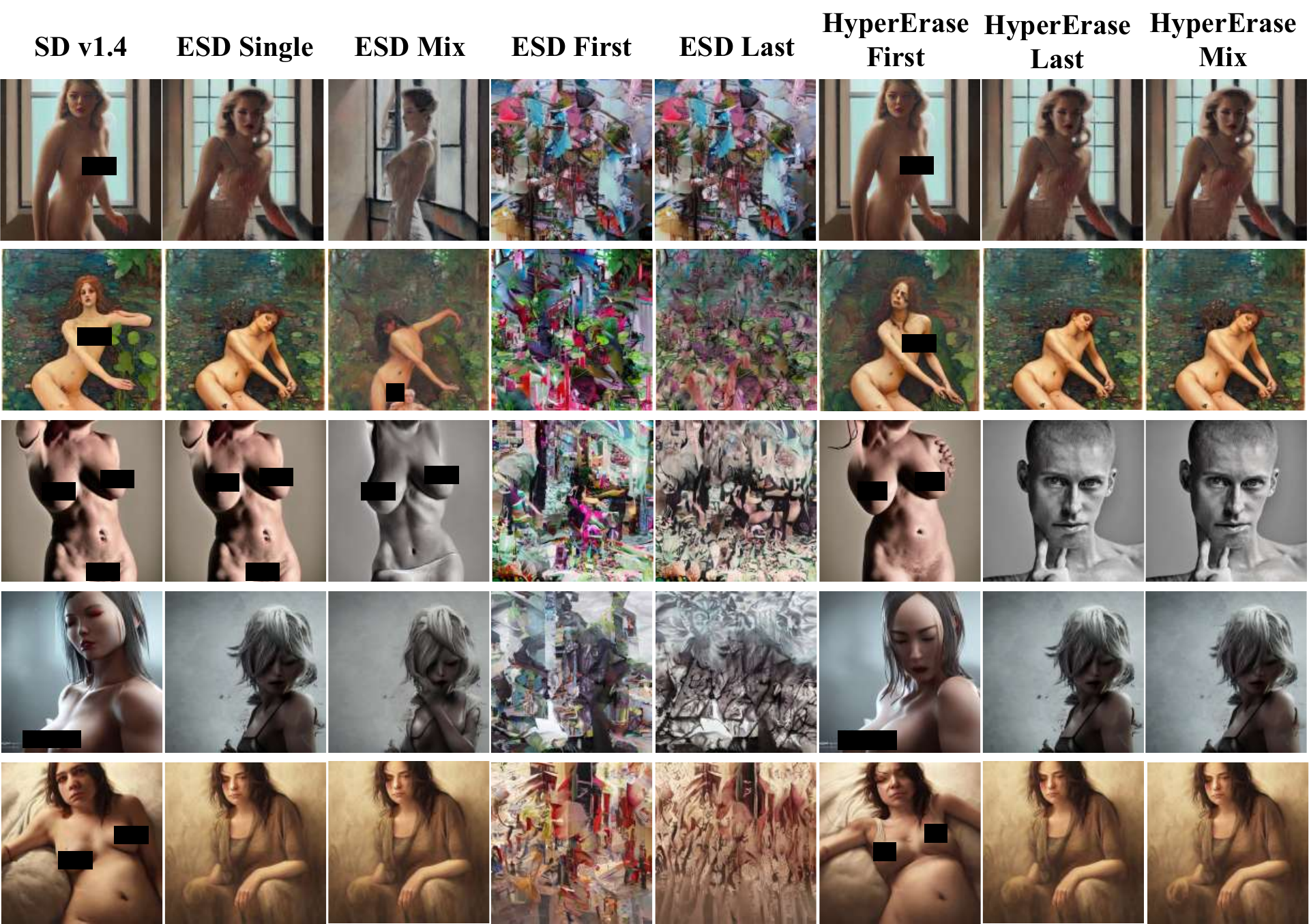}}
  \caption{Additional results of adversarial attacks, including MMA, RAB,P4D and UnlearnDiff.}
 \label{fig:app_nudity}
\end{figure*}

\begin{figure}
  \centering
  \resizebox{\linewidth}{!}{
  \includegraphics{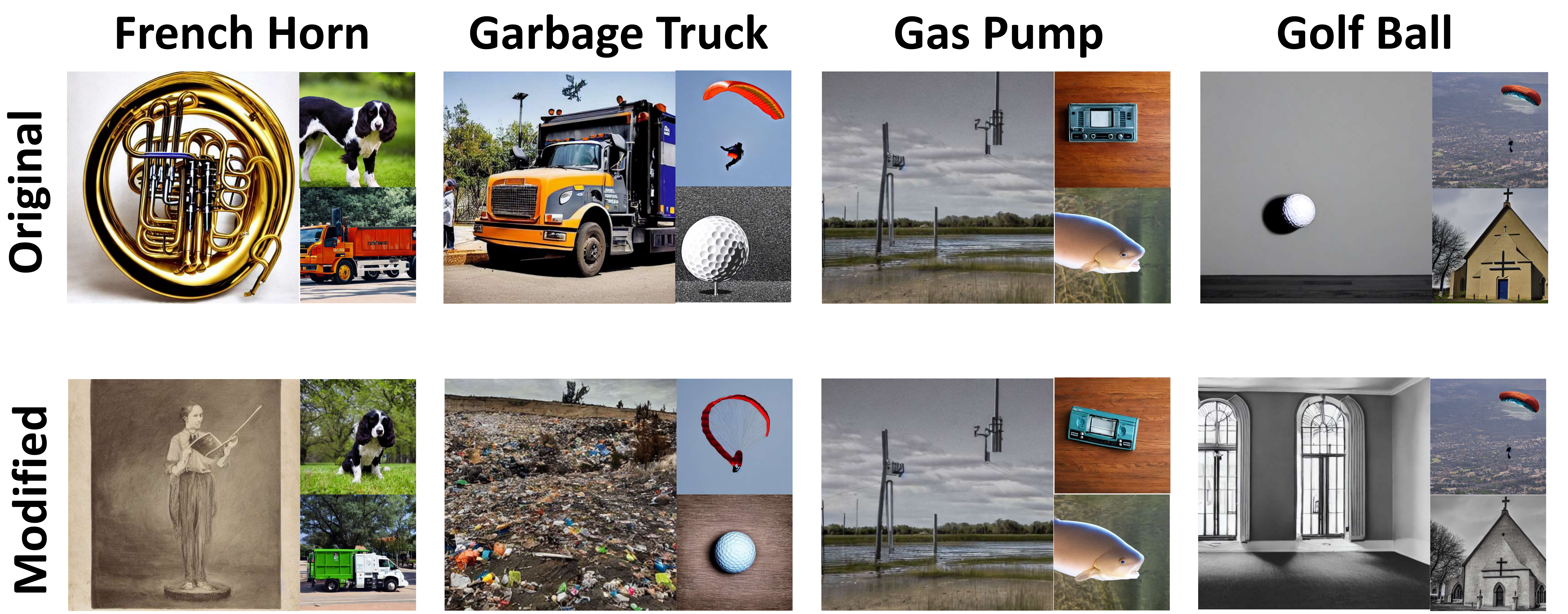}}
  \caption{Additional erasure results of object.}
 \label{fig:app_obj}
\end{figure}

\begin{figure*}
  \centering
  \resizebox{\linewidth}{!}{
  \includegraphics{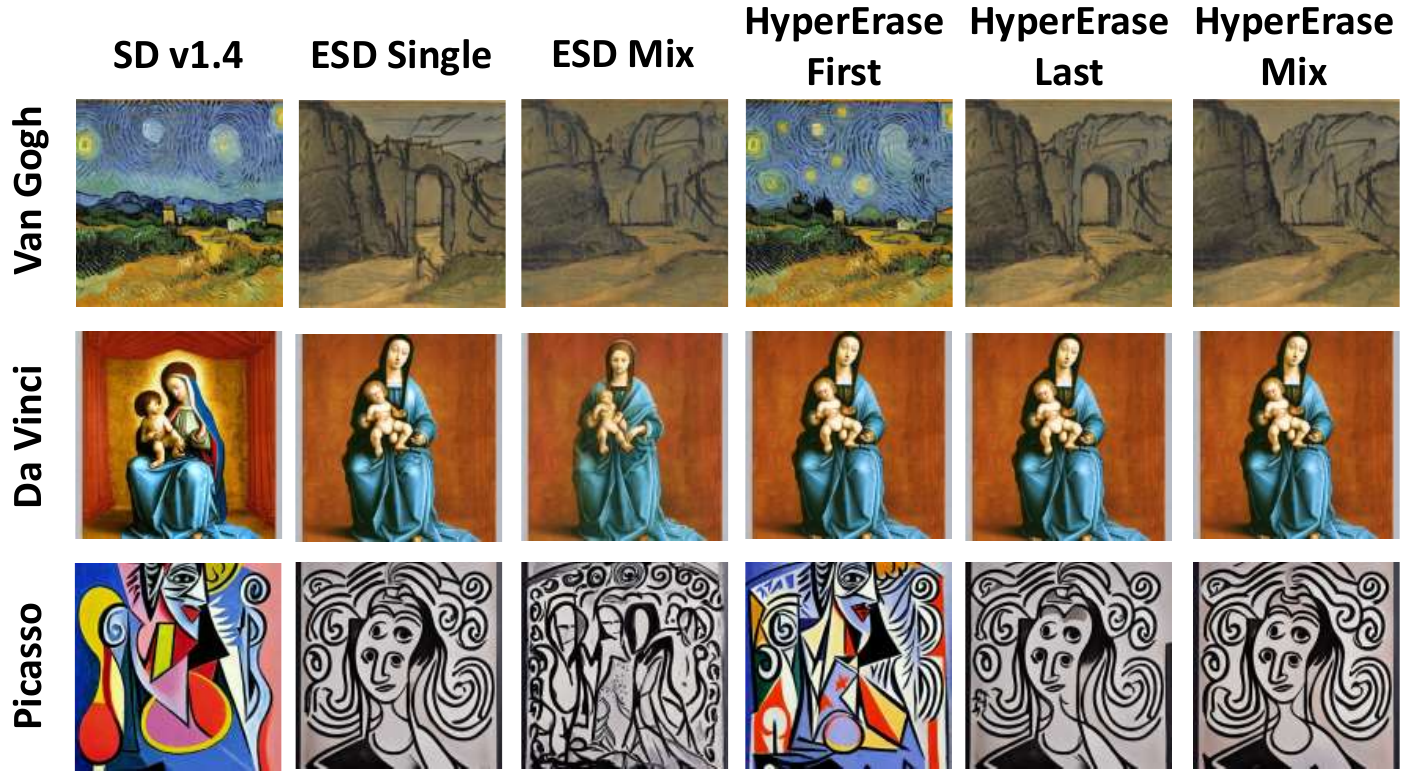}}
  \caption{Additional results of erasing 'Van Gogh'.}
 \label{fig:app_artist}
\end{figure*}

\begin{figure*}
  \centering
  \resizebox{\linewidth}{!}{
  \includegraphics{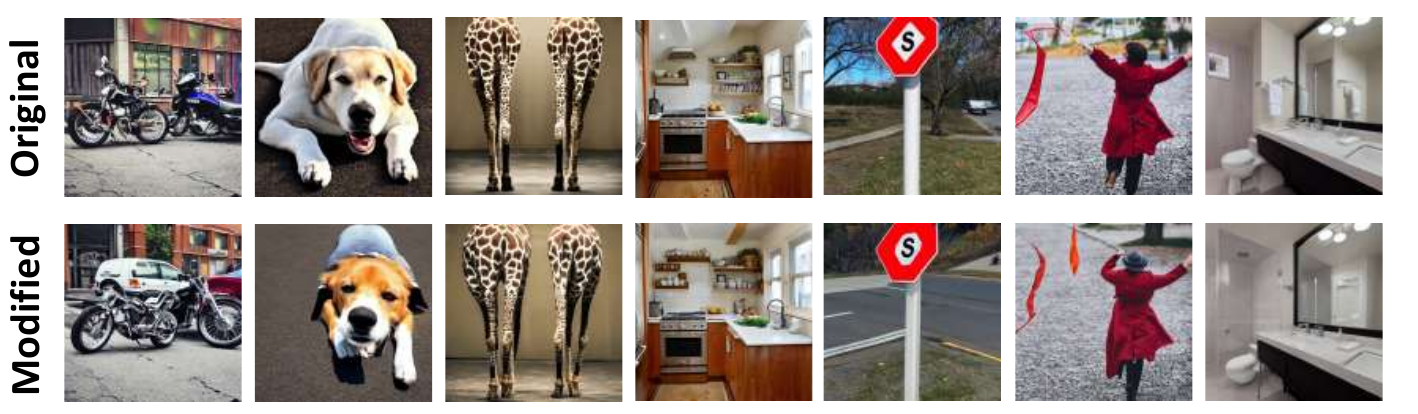}}
  \caption{Comparison of images generated by different methods via MS-COCO dataset.}
 \label{fig:app_coco}
\end{figure*}

In this section we will provide extra visual results. Table~\ref{fig:app_nudity} shows the erasure results for additional Nudity, Table~\ref{fig:app_obj} shows the erasure results for additional Object, and Table~\ref{fig:app_artist} shows the erasure results for additional Artist Style. Table~\ref{fig:app_coco} shows the visual results on the MSCOCO dataset.

\section{Additional Derivations}
\label{add_method}

\subsection{Algorithm}
\label{add_alg}

In this section, we present the algorithm of this paper, as shown in Algorithm~\ref{alg:hypererase}.

\begin{algorithm}[htbp!]
    \caption{HyperErase}
    \label{alg:hypererase}
    \parbox{\linewidth}{
        \textbf{Input}: Tokenized prompt-LoRA training set
        $\mathcal{D}_{\mathrm{PG}}=\{(\mathbf{c}_i,T_i,\mathcal{V}_i)\}$,
        where $\mathbf{c}_i=(c_{i,1},\ldots,c_{i,M})$ is a batch of prompts,
        $T_i\in\mathbb{R}^{N\times h\times t_w}$ is its target token grid,
        and $\mathcal{V}_i$ is the set of valid token positions; frozen BERT
        encoder $E_{\mathrm{BERT}}$; token-importance weights $\{\omega_j\}_{j=1}^{N}$;
        learning rate $\eta$; number of training steps $Q$; fixed scale constant $S$\\
        \textbf{Output}: Trained parameter generator $G_{\psi}$ and a generated
        LoRA adapter $\widehat{\Theta}$ for an input prompt
    }
    \vspace{-2pt}
    \begin{algorithmic}[1]
        \STATE \textbf{Training}
        \STATE Initialize generator parameters $\psi$ and freeze $E_{\mathrm{BERT}}$
        \FOR{$q=1$ \TO $Q$}
            \STATE Sample a minibatch $\{(\mathbf{c}_i,T_i,\mathcal{V}_i)\}_{i=1}^{B}
                   \sim\mathcal{D}_{\mathrm{PG}}$
            \FOR{$i=1$ \TO $B$}
                \STATE $H_i\gets E_{\mathrm{BERT}}(\mathbf{c}_i)$
                \STATE $\widehat{T}_i\gets G_{\psi}(H_i)$
            \ENDFOR
            \STATE $\mathcal{L}(\psi)\gets
                   \dfrac{1}{\sum_{i=1}^{B}|\mathcal{V}_i|}
                   \sum_{i=1}^{B}\sum_{j\in\mathcal{V}_i}
                   \omega_j\bigl(\widehat{T}_{i,j}-T_{i,j}\bigr)^2$
            \STATE $\psi\gets\psi-\eta\nabla_{\psi}\mathcal{L}(\psi)$
        \ENDFOR
        \STATE
        \STATE \textbf{Inference}
        \STATE Input a new prompt $c$ and a reference LoRA structure
               $\Theta_{\mathrm{ref}}$
        \STATE $\mathbf{c}\gets(c,\ldots,c)$ with $M$ repeated prompt slots
        \STATE $\widehat{T}\gets G_{\psi}\!\left(E_{\mathrm{BERT}}(\mathbf{c})\right)$
        \FOR{each token $j=1,\ldots,N$}
            \STATE Extract the normalized pattern $\widehat{\tilde{X}}_j$ from
                   the valid entries of $\widehat{T}[j,1{:}r,:]$
            \STATE Renormalize the pattern:
                   $\widehat{\tilde{X}}_{j,\mathrm{rect}}\gets
                   (\widehat{\tilde{X}}_j-\operatorname{mean}(\widehat{\tilde{X}}_j))\,/
                   \operatorname{std}(\widehat{\tilde{X}}_j)$
            \STATE Read $(\widehat{\mu}_j,\widehat{e}_{\sigma,j})$ from
                   $\widehat{T}[j,-2:,:]$ by averaging its repeated scale entries
            \STATE $\widehat{\sigma}_j\gets
                   S\,\widehat{e}_{\sigma,j}^{\,2}$
            \STATE $\widehat{X}_j\gets
                   \widehat{\tilde{X}}_{j,\mathrm{rect}}\widehat{\sigma}_j+\widehat{\mu}_j$
        \ENDFOR
        \STATE Detokenize $\{\widehat{X}_j\}_{j=1}^{N}$ using
               $\Theta_{\mathrm{ref}}$ to obtain
               $\widehat{\Theta}=\{(\widehat{A}_{\ell},\widehat{B}_{\ell})\}_{\ell=1}^{L}$
        \STATE Apply $\widehat{\Theta}$ to the pretrained U-Net and generate the image
               conditioned on the input prompt
        \RETURN $G_{\psi}$, $\widehat{\Theta}$
    \end{algorithmic}
\end{algorithm}

\subsection{Canonical Scale Anchoring}
\label{app:scale-anchoring}

For concept $k$, collect the raw per-chunk statistics

\begin{equation}
u_i^{(k)}
  = \begin{bmatrix}\mu_i^{(k)}\\ \sigma_i^{(k)}\end{bmatrix},
\qquad
\bar{u}^{(k)}=\frac{1}{S}\sum_{i=1}^{S}u_i^{(k)},
\qquad S=50
\end{equation}

The implementation constructs the canonical target by applying the encoder after the raw-space average

\begin{equation}
u_i^{(k)}\;\longrightarrow\;\bar{u}^{(k)}
\;\longrightarrow\;
T_{\mathrm{scale}}^{(k)}=\operatorname{Enc}(\bar{u}^{(k)})
\end{equation}

Let $\Sigma_k=\operatorname{Cov}_i[u_i^{(k)}]$ be the covariance of the raw scale statistics within concept $k$. Before anchoring, the average squared distance from the concept center is

\begin{equation}
\frac{1}{S}\sum_{i=1}^{S}
\left\|u_i^{(k)}-\bar{u}^{(k)}\right\|_2^2
 =\operatorname{tr}(\Sigma_k)
\end{equation}

After anchoring, every checkpoint associated with $k$ receives the same pre-encoding target $\bar{u}^{(k)}$; hence its within-concept target variance is exactly

\begin{equation}
\operatorname{Cov}_i\!\left[\bar{u}^{(k)}\right]=0,
\qquad
\frac{1}{S}\sum_{i=1}^{S}
\left\|\bar{u}^{(k)}-\bar{u}^{(k)}\right\|_2^2=0
\end{equation}

Consequently, for a fixed concept and the scale-only regression subproblem, the within-concept component of the Bayes risk is removed.  The remaining scale regression problem is \emph{concept-level} rather than checkpoint-level. The generator need only distinguish the canonical
statistics of different concepts. So replacing each sampled scale target by the concept-specific canonical block can increase the precise of variance.

\subsection{Square-root Decoder Bias}
\label{app:sqrt-decoder}

Let the encoded prediction error be

\begin{equation}
\hat{e}_\sigma=e_\sigma+\delta
\end{equation}

Substituting this perturbation into the decoder and subtracting the true standard deviation gives the exact identity

\begin{equation}
\begin{aligned}
\hat{\sigma}-\sigma
 &= S_\sigma\!\left[(e_\sigma+\delta)^2-e_\sigma^2\right] \\
 &= 2S_\sigma e_\sigma\delta + S_\sigma\delta^2 \\
 &= 2\sqrt{S_\sigma\sigma}\,\delta + S_\sigma\delta^2
\end{aligned}
\end{equation}

If $\mathbb{E}[\delta]=0$ and $v_\delta=\mathbb{E}[\delta^2]$, the mean decoded error is

\begin{equation}
\mathbb{E}[\hat{\sigma}-\sigma]=S_\sigma v_\delta
\end{equation}

Thus the square-root decoder converts zero-mean encoded noise into an additive bias proportional to the encoded-error variance and to $S_\sigma$.  For a symmetric error distribution, $\mathbb{E}[\delta^3]=0$, and the leading term of the second moment is linear in $\sigma$

\begin{equation}
\mathbb{E}\!\left[(\hat{\sigma}-\sigma)^2\right]
 \approx 4S_\sigma\sigma\,v_\delta + S_\sigma^2\mathbb{E}[\delta^4]
\end{equation}

For small errors, the relative error has the first-order form

\begin{equation}
\frac{\hat{\sigma}-\sigma}{\sigma}
 = 2\frac{\delta}{e_\sigma} + \frac{\delta^2}{e_\sigma^2}
 \approx 2\frac{\delta}{e_\sigma}
\end{equation}

This equation show that it reduces the source of $\delta$ by removing within-concept scale variation before the generator is optimized, and therefore controls the decoder-induced bias at its source.

\subsection{Inference-time Renormalization}
\label{app:direction-preserving}

Let $P$ denote the valid, non-padding pattern entries of one predicted LoRA
chunk, and let $\mathcal{R}(P)$ denote the $z$-scoring operation implemented during renormalizaion, with entrywise mean $\bar{P}$ and entrywise standard deviation $s_P$.  For any $a>0$ and any constant $b$, the transformed pattern $P'=aP+b\mathbf{1}$ satisfies

\begin{equation}
\overline{P'}=a\bar{P}+b,
\qquad
s_{P'}=a\,s_P,
\end{equation}

and therefore

\begin{equation}
\mathcal{R}(aP+b\mathbf{1})
 =\frac{aP+b\mathbf{1}-(a\bar{P}+b)\mathbf{1}}{a\,s_P}
 =\mathcal{R}(P)
\end{equation}

The operation removes the predicted chunk's affine magnitude and offset while preserving its coordinate-wise orientation for positive scaling. If the detokenized chunk is written as

\begin{equation}
\widehat{X}=\mu\mathbf{1}+\sigma\,\mathcal{R}(P),
\end{equation}

then, provided the same standard-deviation convention is used in both steps,

\begin{equation}
\operatorname{mean}(\widehat{X})=\mu,
\qquad
\operatorname{std}(\widehat{X})=\sigma
\end{equation}

Hence the separately decoded (or canonical) pair $(\mu,\sigma)$ determines the reconstructed chunk magnitude.

For layer $\ell$ and calibration checkpoint $i$, the code measures the effective LoRA update norm

\begin{equation}
q_{\ell,i}=\left\|B_{\ell,i}A_{\ell,i}\right\|_F,
\qquad i=1,\ldots,S,\quad S=50
\end{equation}

The calibration target and the norm of the generated adapter are

\begin{equation}
q_\ell^\star=\frac{1}{S}\sum_{i=1}^{S}q_{\ell,i},
\qquad
\hat{q}_\ell=\left\|\hat{B}_\ell\hat{A}_\ell\right\|_F.
\end{equation}

When $\hat{q}_\ell>10^{-9}$, the implementation uses

\begin{equation}
\gamma_\ell=\frac{q_\ell^\star}{\hat{q}_\ell},
\qquad
\hat{B}_\ell'=\gamma_\ell\hat{B}_\ell,
\qquad
\hat{A}_\ell'=\hat{A}_\ell
\end{equation}

By homogeneity of the Frobenius norm,

\begin{equation}
\left\|\hat{B}_\ell'\hat{A}_\ell'\right\|_F
 =\left\|\gamma_\ell\hat{B}_\ell\hat{A}_\ell\right\|_F
 =\gamma_\ell\hat{q}_\ell
 =q_\ell^\star
\end{equation}

Moreover, the normalized effective update is unchanged

\begin{equation}
\frac{\hat{B}_\ell'\hat{A}_\ell'}
     {\left\|\hat{B}_\ell'\hat{A}_\ell'\right\|_F}
=
\frac{\hat{B}_\ell\hat{A}_\ell}
     {\left\|\hat{B}_\ell\hat{A}_\ell\right\|_F}
\end{equation}

Therefore renormalization changes only the effective layer
magnitude, not the generated LoRA direction or its rank-$r$ subspace.  This explains why calibration can be applied after detokenization without retraining $G_\psi$.

\subsection{First-order LoRA Factor Error}
\label{app:lora-error}

Let the ideal LoRA factors for layer $\ell$ be $A_\ell^\star$ and $B_\ell^\star$, and write the generated factors as

\begin{equation}
\hat{A}_\ell=A_\ell^\star+E_{A,\ell},
\qquad
\hat{B}_\ell=B_\ell^\star+E_{B,\ell}
\end{equation}

The resulting effective update error is exactly

\begin{equation}
\hat{\Delta W}_\ell-\Delta W_\ell^\star
 = B_\ell^\star E_{A,\ell}
   + E_{B,\ell}A_\ell^\star
   + E_{B,\ell}E_{A,\ell}
\end{equation}

Using submultiplicativity of the Frobenius and spectral norms gives

\begin{equation}
\left\|\hat{\Delta W}_\ell-\Delta W_\ell^\star\right\|_F
 \leq
 \left\|B_\ell^\star\right\|_2\left\|E_{A,\ell}\right\|_F
 +\left\|A_\ell^\star\right\|_2\left\|E_{B,\ell}\right\|_F
 +\left\|E_{B,\ell}\right\|_F\left\|E_{A,\ell}\right\|_F
\end{equation}

For small factor errors, the final product term is second order. The first two terms show that errors in the scale rows are amplified by the factor norms, and motivate the decoupling: canonical scale anchoring reduces the
magnitude-related part of $E_{A,\ell}$ and $E_{B,\ell}$, while pattern renormalization controls the standardized-pattern norm before the factors are recoupled. If the generated effective update is approximately a scalar multiple of the ideal direction, $\hat{\Delta W}_\ell\approx a_\ell\Delta W_\ell^\star$, then calibration selects

\begin{equation}
\gamma_\ell\approx
\frac{q_\ell^\star}
     {a_\ell\left\|\Delta W_\ell^\star\right\|_F}
\end{equation}

which cancels the scalar magnitude error while leaving the directional error unchanged.  This is the algebraic separation exploited by the inference-time calibration branch. Scale anchoring controls the source of $\delta$, pattern renormalization controls orientation, and HyperErase calibration corrects residual magnitude without modifying the learned LoRA subspace.

\section{AI Disclosure Section}

In this work, we used generative AI tools for <Generate synthetic data sets, assist in the writing of proofs, assist with translation, create or edit software code, draft parts of a research paper, summarize or analyse existing literature, sourcing/searching for information, edit a research paper to improve readability, identify relevant literature, format references>,. We have not used generative AI tools for <help develop theoretical models or conceptual frameworks, formulate mathematical claims, provide critical ingredients for proving mathematical claims, propose or refine hypotheses, design or provide feedback on research  methodology or experiments, implement methods, interpret results, create or modify scientific figures or images, suggest experimental parameters, discover research topics or identify gaps, brainstorming, suggest a structure for a research paper, propose a title or keywords for a research paper>, and <clean and reformat dataset, support qualitative and thematic data analysis, Formulate questions for surveys or interviews, creation of artifacts, transcribe recordings of research material> are not applicable to this work. Additionally, we used generative AI tools for <tasks with recommended disclosure>. We have reviewed all AI-assisted work. We take responsibility for the final content of this work, including text, claims or artifacts produced with the aid of generative AI.

\end{document}